\documentclass[journal,twoside,web]{ieeecolor}
\usepackage{tmi}
\usepackage{cite}
\usepackage{amsmath,amssymb,amsfonts}
\usepackage{algorithmic}
\usepackage{graphicx}
\usepackage{textcomp}
\usepackage{gensymb}
\usepackage{subfig}
\usepackage{float}
\usepackage{algorithm}
\usepackage{algorithmic}
\usepackage{float}
\usepackage{graphicx}
\usepackage{subfig}
\usepackage{booktabs}
\usepackage{multirow}
\usepackage{cleveref}
\usepackage{amssymb}
\usepackage{pifont}
\usepackage{color}
\usepackage[table]{xcolor}
\usepackage{tabu}
\usepackage{cleveref}
\usepackage{makecell}
\usepackage{placeins}

\def\BibTeX{{\rm B\kern-.05em{\sc i\kern-.025em b}\kern-.08em
    T\kern-.1667em\lower.7ex\hbox{E}\kern-.125emX}}
\begin{document}
\bstctlcite{IEEEexample:BSTcontrol}
\title{Clinical Metadata Guided Limited-Angle CT Image Reconstruction}
\author{Yu Shi, \IEEEmembership{Member, IEEE}, Shuyi Fan, Changsheng Fang, Shuo Han, Haodong Li, Li Zhou, Bahareh Morovati, Dayang Wang, \IEEEmembership{Senior Member, IEEE}, and Hengyong Yu, \IEEEmembership{Fellow, IEEE}
\thanks{The paper was submitted on XXXXX. This work was supported in part by NIH/NIBIB under grants R01EB032807 and R01EB034737, and NIH/NCI under grant R21CA264772. Corresponding author: H. Yu, Email: hengyong-yu@ieee.org.}
\thanks{The authors are with the Department of Electrical and Computer Engineering, University of Massachusetts Lowell, Lowell, MA, 01854, USA.}
}

\maketitle

\begin{abstract}
Limited-angle computed tomography (LACT) improves temporal resolution and reduces radiation dose, but suffers from severe artifacts due to missing projections. Clinical workflows record abundant patient- and acquisition-level metadata, yet such information remains underutilized in image reconstruction. To tackle the ill-posed LACT inverse problem, we propose a metadata-guided two-stage diffusion framework that leverages structured clinical contexts as semantic priors for robust reconstruction. In Stage-I, we learn a metadata-to-anatomy generative prior by conditioning a transformer-based diffusion model on clinical metadata (acquisition parameters, patient demographics, and diagnostic impressions), and sampling a coarse anatomical estimate from Gaussian noise. In Stage-II, a second conditional diffusion model performs coarse-to-fine refinement, using the Stage-I estimate as an image prior while re-injecting the same metadata to recover full-resolution anatomy. To preserve anatomical fidelity and suppress hallucinations, projection-domain data consistency is enforced periodically after denoising update via an ADMM-based solver.
Experiments on the public multimodal CTRATE dataset demonstrate that the proposed framework outperforms iterative, CNN-based, and diffusion-based baselines, with the greatest gains under severe truncation, \emph{e.g.}, up to 5.23\%/11.21\% higher SSIM/PSNR than the strongest metadata-free diffusion competitor at 90{\degree}. On real clinical cardiac CT, it yields coronary artery calcium scores closer to full-view references, indicating improved clinical utility. Furthermore, the proposed method is generalized to out-of-distribution angular ranges and projection geometries, and ablation results confirm complementary contributions from different metadata types under limited-angle conditions. Our results highlight clinical metadata as actionable semantic priors to synergize with physics-informed constraints to improve both reconstruction fidelity and clinical quantification in LACT.
\end{abstract}


\section{Introduction}
\label{sec:introduction}
\IEEEPARstart{C}{omputed} {tomography (CT) is a cornerstone of medical imaging, supporting chest and cardiac screening\cite{lubbers2018comprehensive, national2011reduced, de2020reduced}, preoperative planning\cite{he2025three}, intraoperative navigation\cite{key2023cone}, and treatment response assessment\cite{Kashyap_2025}. High-fidelity reconstruction typically requires sufficient angular coverage to ensure stable inversion and near-isotropic spatial resolution. In practice, such coverage is not always available due to physical and/or clinical constraints. First, acquisition trajectories may be restricted by system kinematics and workflow. For example, in interventional suites using C-arm cone-beam CT, table layout and instrument collisions can preclude a full circular orbit, motivating noncircular trajectories or short-arc acquisitions\cite{LEE2024169393}. 
Second, for dynamic organs, improving temporal resolution requires reducing the data-acquisition window. In cardiac CT, motion artifacts may persist even with fast gantry rotation (\emph{e.g.} 230 ms per 360\textdegree) while coronary assessment often benefits from sub-100-ms temporal resolution\cite{toia2020technical,mergen2023importance,sartoretti2025effect}. Reducing the scanning angle provides a direct way to accelerate acquisition\cite{Zhou2021limited}, because the temporal window scales approximately with angular span (\emph{e.g.} 57.5 ms for 90{\textdegree} at 230 ms/360\textdegree), potentially improving motion robustness without hardware changes. Meanwhile, reducing angular coverage also reduces total mAs approximately proportionally and thus radiation exposure.
In this context, limited-angle computed tomography (LACT) has emerged as a promising solution to these challenges in clinical practice\cite{apfaltrer2013enhanced}.} However, the resulting angular truncation renders reconstruction severely ill-posed, often introducing pronounced streak artifacts, structural blurring, and intensity bias, especially near anatomical boundaries\cite{bappy2025deep}.

Classical analytical methods and iterative reconstruction with handcrafted priors \cite{ZHANG2021102030, WANG2023128013} struggle under severe angular truncation, often failing to recover fine anatomical structures or suppress artifacts \cite{huang2018scale, xu2024hybrid}. Recent advances in deep learning have significantly improved LACT reconstruction by learning data-driven priors from large-scale datasets. Early convolutional neural network (CNN)-based approaches, such as FBPConvNet\cite{Jin2017}, demonstrated remarkable capability in restoring missing structures from undersampled projections. However, their limited receptive field and poor adaptability across different scan protocols and anatomies hinder broader deployment. In contrast, Transformer architectures address these limitations by capturing long-range contextual features via global attention mechanisms instead of localized convolutions. Recent studies\cite{pan2022multi,xu2024hybrid,Bahareh2025} explored attention-based architectures to improve CT reconstruction under severe noise or data sparsity, yet most existing models rely solely on image-based supervision and overlook valuable clinical metadata that routinely accompany imaging exams. The disconnect between imaging data and associated metadata suggests an untapped potential for metadata-aware reconstruction frameworks to enhance image fidelity in ill-posed settings.

With the increasing availability of multimodal datasets that pair medical images with textual records, there is growing recognition that image reconstruction should not be treated as a purely image-domain task. In particular, Transformer-based models naturally support the fusion of image features and non-image information via cross-attention, enabling semantic conditioning beyond pixel supervision. {Motivated by this capability, several recent studies\cite{ma2023prompted, ma2025tmi, li2024progressively, shi2026prompt} explored prompt-conditioned reconstruction, \emph{e.g.}, prompt-guided models for sparse-view CT reconstruction and low-dose metal artifact reduction, highlighting the emerging synergy between text conditioning and medical imaging.}
However, {most approaches mainly focus on protocol-level descriptors or feature-derived prompts, which may capture only coarse acquisition characteristics and underutilize richer, patient-specific clinical metadata}, thereby limiting the semantic context available for guidance. In routine clinical practice, metadata are inherently diverse\cite{gauriau2020using}, spanning acquisition parameters, patient demographics, and diagnostic impressions. These heterogeneous sources provide complementary constraints that can jointly narrow the feasible solution space in ill-posed reconstruction.

To incorporate rich semantic conditioning into learned priors, recent studies integrated Transformer components into diffusion models, opening new frontiers in generative medical imaging\cite{bao2023all,ma2024efficient}.
As probabilistic generators, diffusion models are well-suited for modeling the underlying distribution of high-dimensional medical data and learning expressive priors. With appropriate conditioning, these priors can be steered by semantic cues during sampling. Although the application of text-conditioned diffusion models to inverse imaging problems remains nascent, early evidence suggests improved image quality under incomplete measurements. For example, ContextMRI\cite{chung2025contextmri} conditions a diffusion model for compressed-sensing MRI on detailed clinical context (anatomical region, contrast, acquisition parameters, and pathology descriptions) via CLIP-based embeddings, outperforming unconditional counterparts under extreme undersampling at a potential cost of reduced generalizability. In the field of CT image synthesis, GenerateCT\cite{Hamamci2024GenerateCT} employs a vision-language transformer to synthesize low-resolution CT images from text and refines it with a diffusion model.
These approaches collectively suggest a promising direction for ill-posed image reconstruction by combining diffusion priors with multimodal clinical conditioning.

\begin{figure}[!htb]
\centerline{\includegraphics[width=3.5in]{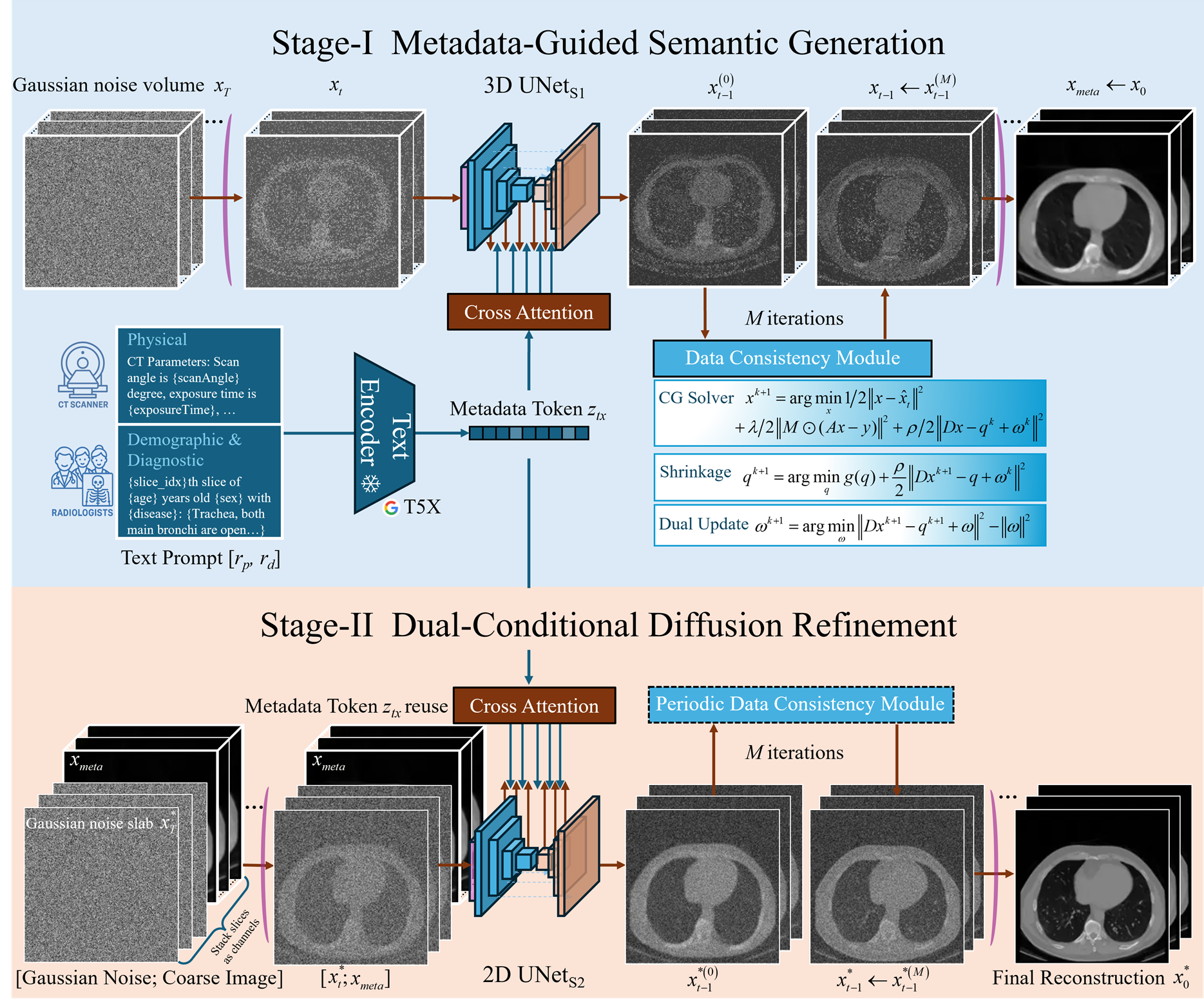}}
\caption{Overview of the proposed two-stage, metadata-guided diffusion framework for LACT reconstruction. 
}
\label{fig1}
\end{figure}

Despite the growing success of generative modeling in medical imaging, LACT reconstruction remains substantially more challenging than conventional image generation or super-resolution. 
First, medical reconstruction demands strict anatomical fidelity, leaving little tolerance for artificial lesions or structural hallucinations, which imposes stringent constraints on the generative process.
Second, the image degradation caused by limited-angle scanning settings poses an unfavorable condition for diffusion models, making the denoising trajectory more difficult to converge.
Furthermore, diffusion-based LACT reconstruction that leverages structured clinical metadata as semantic conditioning remains underexplored.
To address these challenges, we propose a metadata-guided two-stage diffusion framework for LACT reconstruction, as shown in Fig.\ref{fig1}. The framework progressively synthesizes and refines CT images from clinical metadata while enforcing projection-domain consistency with the measured incomplete sinograms, thereby obtaining high-fidelity CT images suitable for {clinical applications}.

Main contributions of this work are summarized as follows:
\begin{itemize}
\item[$\bullet$]Metadata-guided two-stage framework: A cascaded diffusion pipeline is proposed for LACT reconstruction implemented in a lightweight 2.5D slab setting. Stage-I synthesizes coarse anatomical priors, and Stage-II performs dual-conditioned refinement using both the Stage-I prior and the same metadata embeddings.

\item[$\bullet$] Heterogeneous clinical metadata as semantic priors: We unify acquisition parameters, demographics, and diagnostic impressions into a single conditioning interface for both stages, enabling image-free semantic prior synthesis in Stage-I and complementary guidance in Stage-II.

\item[$\bullet$] Stage-synchronized supervised fine-tuning (SFT): Beyond separate pretraining of the two stages, we introduce an additional supervised fine-tuning phase that jointly optimizes both stages end-to-end, improving cross-stage compatibility and mitigating error accumulation from Stage-I to Stage-II.

\item[$\bullet$] {Physics-informed sampling with projection-domain data consistency: We periodically interleave diffusion denoising with an alternating direction method of multipliers (ADMM)-based data-consistency solver} to ensure alignment between semantic priors and measurements.

\item[$\bullet$] Comprehensive validation with robustness and clinical relevance: Extensive experiments on the public multimodal dataset and real cardiac LACT show consistent gains over strong baselines, including generalization to unseen angular ranges and projection geometries, ablation-verified complementary contributions of different metadata types, and more accurate CAC risk stratification relative to full-view references.
\end{itemize}


\section{Methodology}
\label{sec:methodology}
\subsection{Problem Formulation and Framework Overview}
The CT imaging process can be modeled as a discrete linear system:
\begin{equation}y=Ax+b,\label{eq1}\end{equation}
where $x\in\mathbb{R}^{H_{im}W_{im}}$ denotes the vectorized object to be reconstructed with height $H_{im}$ and width $W_{im}$, $y\in\mathbb{R}^{P_{pj}}$ is the measured vectorized sinogram with a total measurement of $P_{pj}$, $A\in\mathbb{R}^{P_{pj}\times H_{im}W_{im}}$ is the system matrix representing the forward projection operator, and $b\in\mathbb{R}^{P_{pj}}$ represents measurement noise. {For clarity, we present the forward model in 2D, and the same formulation extends to 3D geometry by replacing the projector/backprojector with their 3D counterparts and redefining variable dimensions.} In LACT, only a subset of projection angles is available. Let $M\in\{0,1\}^{P_{pj}}$ denote a binary angular truncation mask that selects the acquired views, then the measurement model is reformulated as (\ref{eq2}):
\begin{equation}
y=M\odot(Ax+b),
\label{eq2}
\end{equation}
where $\odot$ denotes element-wise multiplication. This angular truncation yields incomplete projection, creating a severely ill-posed inverse problem where conventional analytical methods fail to recover fine structures.

To address this LACT reconstruction problem, we propose a generative reconstruction framework that integrates semantic priors derived from structured clinical metadata with projection-domain constraints. As shown in Fig.\ref{fig1}, the pipeline consists of two cascaded diffusion stages.
Stage-I learns a conditional diffusion prior that synthesizes a coarse anatomical estimate from pure Gaussian noise using only clinical metadata, including structured scan parameters (\emph{e.g.}, exposure time, scan angle), demographics (\emph{e.g.}, age, sex) and radiological impressions. Metadata are encoded with a pretrained T5X encoder\cite{raffel2020exploring,roberts2022t5x}, and injected into a transformer-conditioned UNet via cross-attention to steer the denoising trajectory under the Elucidated Diffusion Model (EDM) sampling formulation. To prevent unconstrained hallucination and to anchor the coarse estimate to the measured sinogram, Stage-I applies a projection-domain data consistency (DC) correction after each denoising update, implemented with an ADMM-based solver.

Stage-II performs coarse-to-fine reconstruction at higher spatial resolution by conditioning on both the Stage-I estimate as an image prior and the same metadata embeddings. This stage focuses on recovering high-frequency details and improving local anatomical consistency while still respecting the measured projections. For efficiency and to balance prior-driven completion with physics fidelity, Stage-II interleaves EDM sampling with the same ADMM-based DC module periodically rather than at every timestep. Overall, the cascaded design progressively narrows the feasible solution space using metadata-derived semantics, while the DC module enforces measurement consistency throughout sampling, yielding reconstructions that are both anatomically meaningful and physically plausible.

\subsection{Data Curation}
\label{sec:data_curation}
The training dataset is derived from the CTRATE dataset\cite{hamamci2024developing}, a public collection of chest CT volumes with paired acquisition parameters, demographics, and diagnostic reports, making it well-suited for multimodal learning.
From the CTRATE training split, we randomly select 1,000 patient cases. {To reduce computational cost while preserving short-range through-plane context, we adopt a slab-based 2.5D representation by extracting overlapping slabs of nine contiguous axial slices with stride 1 from each 3D volume. Each training instance consists of a nine-slice full-angle CT slab and its corresponding limited-angle measurements. The reconstruction target is defined as the central three slices of the slab.}

Since clinical text metadata are inherently scan-level, the original radiology report is shared across all slabs from the same CT volume. To provide additional structured diagnostic cues, we follow CTRATE’s protocol and use CT-CLIP\cite{hamamci2024developing}, a language-image pretraining framework for CTRATE, to predict diagnostic labels from slabs of nine contiguous axial slices. {We then retain all slabs with at least one predicted abnormality. For volumes in which CT-CLIP predicts no abnormalities across all slabs, we randomly retain one slab to represent the normal case}. Additional metadata, including patient age, sex, and axial position within the 3D volume, are also included for each instance. Inspired by recent success in metadata-guided reconstruction for MRI\cite{chung2025contextmri} and CT\cite{ma2023prompted}, which demonstrate the importance of acquisition parameters in learning-based recovery from sparse measurements, we also incorporate acquisition parameters as part of metadata. To standardize the metadata conditioning interface across models, all available metadata are formatted into structured natural-language prompts following CT-CLIP conventions, as summarized in Table \ref{tab1}.
\begin{table}[H]
\caption{Example of Metadata Input}
\label{tab1}
\centering
\setlength{\tabcolsep}{3pt}
\begin{tabular}{|p{45pt}|p{185pt}|}
\hline
Type&
Input Content\\
\hline
Physics& 
CT Parameters: Scan angle is \{scanAngle\} degree, exposure time is \{exposureTime\}, X-Ray tube current is \{tubeCurrent\}, exposure is \{exposure\} mAs.\\
Demography \& Diagnosis& 
Slices \{slice range\} of \{age\} years old \{sex\} with \{diseases\}: \{Trachea open, left lung partially collapsed, large pleural effusion...\}. \\
\hline
\end{tabular}
\end{table}

In total, the training dataset comprises 4,625 full-angle slabs. For each slab, we forward-project each axial slice using CTLIB\cite{xia2021magic} to generate a stack of fan-beam sinograms {under the following geometry: the reconstruction grid is 512$\times$512 with an in-plane pixel size of 0.69 mm, the source-to-isocenter distance (SID) is 550 mm, the source-to-detector distance (SDD) is 950 mm, the detector is equi-angular with 900 elements, and the angular pitch per element is 1/950 rad ($\sim0.06\degree$), corresponding to an effective arc length of 1.00 mm at the detector. We uniformly sample 1,000 projection views over 360${\degree}$. Limited-angle data are generated by angular truncation to $60{\degree}$, 90${\degree}$, 120${\degree}$, in addition to the 360${\degree}$ full-angle case, yielding a total of 18,500 instances (slab, sinogram stack, metadata). The ground-truth slices have compact support on the $512\times512$ grid with no unknown anatomy outside the field of view. To simulate different dose/noise levels, we adopt a standard transmission CT measurement model\cite{Liu_2012}, where the incident photon count is scaled proportionally to mAs, transmitted photon counts follow Poisson statistics optionally with additive Gaussian electronic noise, and noisy line integrals are obtained via the log transformation.}

For testing, we randomly sample 200 patients from the CTRATE testing split and apply the same preprocessing and slab extraction pipeline, yielding 3,700 CT slab-sinogram stack-metadata pairs. Besides CTRATE, we evaluate generalizability on a clinical cardiac dataset acquired on a GE Discovery CT750 HD scanner, approved by the Institutional Review Boards of the Vanderbilt University Medical Center and the University of Massachusetts Lowell. {The full-angle clinical scans are acquired using prospectively electrocardiogram (ECG)-gated axial acquisition (GE SnapShot Pulse), obtaining a single cardiac phase with minimal motion artifacts.} A total of {21} patient raw projection data are acquired over a full angular range of 360${\degree}$, yielding 984 views per case, with each projection consisting of 821 detector elements. The SID and SDD are 625.6 mm and 1097.0 mm, respectively. {To keep the metadata conditioning format consistent with CTRATE, we use CT-CHAT (a CT-CLIP–based vision-language chat model) to synthesize a scan-level diagnostic impression and slab-level disease labels, and cast them into the same prompt template used for CTRATE. In contrast, acquisition parameters and demographic attributes are obtained from the scanner records and treated as real metadata. We use the corresponding full-angle filtered backprojection (FBP) reconstructions as the references for clinical evaluation, and simulate limited-angle scans by angular truncation of the same raw sinograms at 90${\degree}$ and 120${\degree}$}.

\subsection{Metadata-guided Diffusion for LACT Generation}
\subsubsection{Metadata-conditioned Diffusion Sampling}
The first stage employs a weakly conditional diffusion model to generate a coarse CT image directly from clinical metadata. Unlike supervised methods that condition on corrupted reconstructions (\emph{e.g.} FBP images), Stage-I synthesizes the anatomical structures from noise while being guided by semantic priors encoded in textual metadata.
Specifically, each slab instance is paired with a structured text prompt $r=[r_p;r_{de};r_{di}]$, where $r_p$ contains physical acquisition parameters, $r_{de}$ includes patient demographic information, and $r_{di}$ represents diagnostic impressions constructed by aligning the scan-level report with the corresponding slab context. These textual inputs are embedded into a token sequence $z_\text{tx}\in\mathbb{R}^{N_t\times {d_\text{emb}}}$ using a pretrained language encoder T5X, where $N_t$ is the number of text tokens and {$d_\text{emb}$} is the embedding dimension.

The image generation process begins from a random noise image $x_T\sim N(0,I)$, where $T$ is the maximum continuous time/noise level. A model parameterized via $\theta_1$ iteratively denoises the image across a sequence of diffusion steps $t=T\rightarrow0$, following the velocity-based EDM formulation:
\begin{equation}
\begin{split}
x_{t_{i-1}}&=x_{t_i}-\Delta t_i\cdot d_i +\sqrt{\omega(t_i)\Delta t_i}\cdot \epsilon_i\ ,\text{with} \\ 
d_i&=v_{\theta_1}(x_{t_i},t_i,z_{\text{tx}})+\frac{1}{2}\omega(t_i)\cdot s_{\theta_1}(x_{t_i},t_i,z_{\text{tx}}),\\
\epsilon_i&\overset{i.i.d.}{\sim} \mathcal{N}(0,I),
\end{split}
\label{eq3}
\end{equation}
where $v_{\theta_1}$ denotes the predicted velocity field, $s_{\theta_1}$ is the approximated score function, representing the log-density gradient at time $t$, and $\omega(t)$ denotes a time-dependent noise scale. {We consider a continuous-time diffusion process indexed by $t$ and discretize it into a sequence $t_i$, where $i$ indexes the discrete diffusion timestep, and $\Delta t_i=t_i-t_{i-1}>0$ denotes the step size between adjacent steps.} The final output of Stage-I at $t_0=0$, denoted $x_\text{meta}$, serves as a coarse anatomical condition for Stage-II. {In our slab-based 2.5D setting, Stage-I operates on a noisy slab $x_t$ and employs a 3D UNet backbone to preserve short-range through-plane contexts.}

Stage-II performs dual-conditional refinement to recover full-resolution anatomy. It conditions on both the image-level features from coarse ${x}_\text{meta}$ and the metadata embedding $z_\text{tx}$, forming a dual-conditioning mechanism that improves detail recovery and suppresses residual artifacts.
At each sampling timestep $t_i$, the model receives the noisy image $x_{t_i}$ and produces an intermediate estimate via the update:
\begin{equation}
\begin{split}
x^{*}_{t_{i-1}}=&x^{*}_{t_i}-\Delta t_i\cdot d^{*}_i +\sqrt{\omega(t_i)\Delta t_i}\cdot \epsilon_i^{*},\text{with} ~\epsilon^*_i\overset{i.i.d.}{\sim} \mathcal{N}(0,I)\\ 
d^{*}_i=&v_{\theta_2}(x^*_{t_i},t_i,x_{\text{meta}},z_{\text{tx}})+\frac{1}{2}\omega(t_i)\cdot s_{\theta_2}(x^*_{t_i},t_i,x_{\text{meta}},z_{\text{tx}}).
\end{split}
\label{eq4}
\end{equation}
Here, $\theta_2$ denotes the parameters of the Stage-II diffusion model, and $x^*_{t_i}$ denotes the Stage-II sampling trajectory conditioned on $[x_{\text{meta}}; z_\text{tx}]$. {For efficient 2.5D implementation, Stage-II uses a 2D UNet backbone and represents the slab depth as channels.} Both stages are trained using pixel-wise $L_2$ loss between the predicted output and the corresponding reference.

\subsubsection{Model Architecture}
To incorporate metadata prior, both stages adopt a cross-attention-augmented UNet that couples convolutional feature extraction with transformer-style conditioning. The noisy image input, either $x_t$ in Stage-I or the cascaded pair $[x^*_t;x_\text{meta}]$ in Stage-II, is first mapped into spatial feature tokens $z_\text{im}\in\mathbb{R}^{N_v\times d_\text{emb}}$ via a multi-scale patch-embedding module. At each encoder/decoder level $l$, the image tokens $z^l_\text{im}$, the metadata tokens $z_\text{tx}$, and a continuous-time embedding $h_{time}$ are jointly provided to the core encoder/decoder blocks.

\begin{figure*}[!htb]
\centering
\centerline{\includegraphics[width=6.0in]{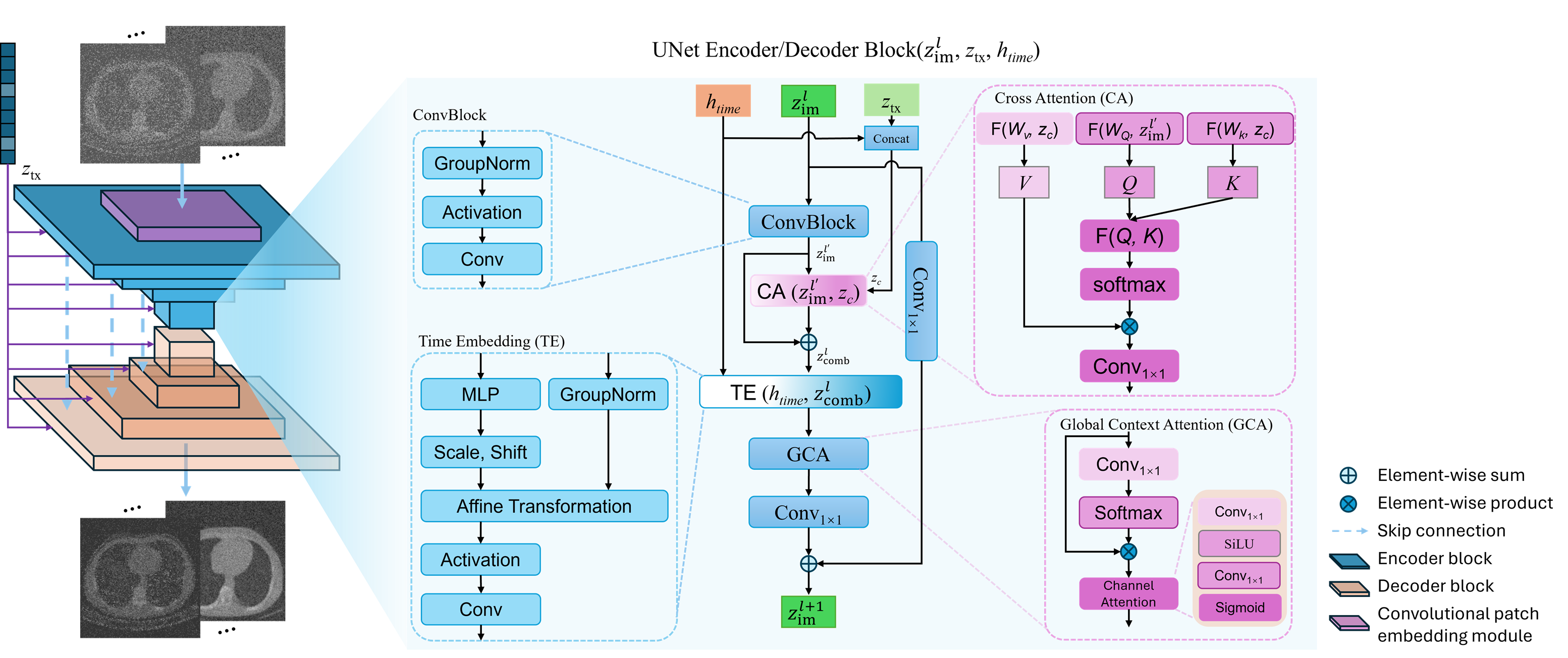}}
\caption{Internal architecture of the conditional UNet Encoder/Decoder Block in the proposed diffusion framework.}
\label{fig2}
\vspace{-3mm}
\end{figure*}

As detailed in Fig.\ref{fig2}, each encoder/decoder block comprises convolutional block (ConvBlock), metadata-aware cross-attention (CA), time embedding (TE) modulation, and guided contextual attention (GCA). \textbf{ConvBlock} first extracts locally processed features $z_\text{im}^{l'}$ given image tokens $z^l_\text{im}$. We form an augmented conditioning token sequence $z_c=\text{Concat}(z_\text{tx},h_{time})$, where $h_{time}\in \mathbb{R}^{1\times d_\text{emb}}$ is projected to the same embedding dimension as $z_\text{tx}$ and appended as an additional token. \textbf{CA} module injects metadata-dependent semantics into the generation process by attending $z^{l'}_\text{im}$ to conditioning tokens $z_c$ as:
\begin{equation}
Q=z_{\text{im}}^{l'}W_Q,~K=z_cW_K,~V=z_cW_V,
\label{eq5}
\end{equation}
\begin{equation}
\text{CA}(z_\text{im}^{l'},z_c)=(\text{softmax}(\frac{QK^{\top}}{\sqrt{d_h}})V)W_O,
\label{eq6}
\end{equation}
where $d_h$ denotes the attention head dimension, $W_Q$, $W_K$, $W_V\in\mathbb{R}^{d_\text{emb}\times Hd_h}$, and $W_O\in\mathbb{R}^{Hd_h\times d_\text{emb}}$ are learnable projection matrices. $Q,K,V$ are reshaped into $H$ heads of dimension $d_h$ and the softmax is applied within each head. Next, \textbf{TE} encodes the continuous noise scale using log-SNR, which is first embedded via sinusoidal positional encoding, and then projected by a multilayer perceptron (MLP) into FiLM-style scale and shift parameters. These parameters modulate GroupNorm-normalized feature maps via an affine transformation, enabling time-aware adaptation of the feature stream.
The modulated features are subsequently passed through an activation function and a convolution layer, providing noise-level-dependent conditioning, aligned with the conditioning paradigm in EDM.
Finally, \textbf{GCA} is a spatial feature matching mechanism to employ channel attention to further enhance structural coherence, and is inserted after the CA module and TE modulation. The updated feature is computed as:
\begin{equation}
\left\{
\begin{aligned}
z_\text{im}^{l'}&=\text{ConvBlock}(z_\text{im}^l)\\
z_\text{comb}^{l}&=z_\text{im}^{l'}+\text{CA}(z_\text{im}^{l'},z_c)\\
z_\text{im}^{l+1}&=\text{Conv}(z_\text{im}^l)+\text{Conv}(\text{GCA}(\text{TE}(h_{time},z_\text{comb}^{l}))).
\end{aligned}
\right.
\label{eq7}
\end{equation}
The output is propagated to the next UNet layer as $z_\text{im}^{l+1}$. All convolutional operators correspond to 3D operators in Stage-I and 2D operators in Stage-II.

\subsubsection{Data Consistency Regularization}
\label{SecDataConsis}
\begin{figure}[!htb]
\centerline{\includegraphics[width=3.0in]{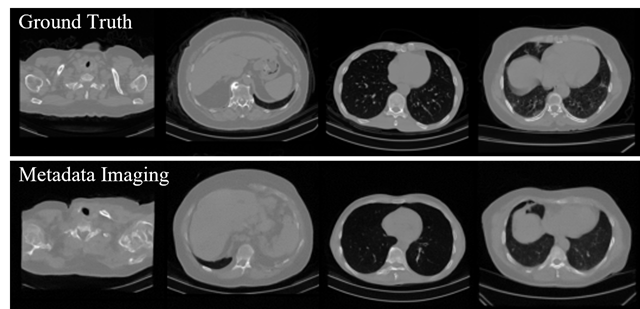}}
\caption{Visual comparison of metadata-conditioned image generation with ground-truth chest CT cases. Each column pair shows the synthesized image from the metadata-guided diffusion model (bottom) and the full-view reference (top). 
}
\label{fig3}
\vspace{-3mm}
\end{figure}

Although the metadata-conditioned diffusion model provides informative semantic priors, the generated image at each timestep may still deviate from the measurement-consistent solution space due to the absence of explicit alignment with the measured sinogram, as shown in Fig.\ref{fig3}. To bridge this gap, we integrate an ADMM-based data consistency module into the sampling loop. Specifically, given the network prediction $\hat{x}_{t_{i-1}}$ after a denoising update, the corrected image $x_{t_{i-1}}$ is computed by solving:
\begin{equation}
\begin{split}
x&_{t_{i-1}}=\text{arg}\min_x\frac{1}{2}\|x-\hat{x}_{t_{i-1}}\|^2\\
+&\frac{\lambda}{2}\|M\odot(Ax-y)\|^2+\mu R(Dx),
\end{split}
\label{eq8}
\end{equation}
where $A$ is the forward projector, $M$ is the angular mask as defined in Sec. II-A, and $y$ is the measured LACT sinogram. $R(Dx)$ denotes isotropic total variation (TV) regularizer with $D$ the discrete gradient operator. The weights $\lambda$ and $\mu$ are hyperparameters to balance data fidelity and regularization.

To solve this problem efficiently, we introduce an auxiliary variable $q=Dx$ and recast the regularization as a separable constraint. The corresponding augmented Lagrangian is:
\begin{equation}
L_p(x,q,\omega)=f(x)+g(q)+\frac{\rho}{2}\|Dx-q+\omega\|^2-\frac{\rho}{2}\|\omega\|^2,
\label{eq10}
\end{equation}
where $f(x)$ collects all the $x$-related terms from the network proximity and data-fidelity penalties, $g(q)$ denotes TV regularization, $\omega$ is the scaled dual variable, and $\rho$ is the penalty parameter. The iterative ADMM is applied to solve Eq.(\ref{eq10}). In practice, the resulting x-subproblem is a quadratic least-squares problem, {efficiently solved by conjugate gradient (CG)} method\cite{chung2023decomposed}. The q-subproblem admits a closed-form vector shrinkage update for isotropic TV, followed by a closed-form dual update for $\omega$. We run a small fixed number of ADMM iterations per DC call. This correction ensures the denoised image at each diffusion sampling step remains both semantically plausible and physically consistent. {Consistent with Fig.\ref{fig1}, the DC module is applied at every diffusion step in Stage-I, while it is interleaved periodically during Stage-II refinement to reduce computation.} {During inference, the forward projection operator in the ADMM module is configured to match the corresponding scanner geometry.}

\subsubsection{{Supervised Fine-Tuning (SFT)}}
{To mitigate cross-stage mismatch and error accumulation, we introduce an additional supervised fine-tuning phase that jointly optimizes Stage-I and Stage-II. Starting from the independently pretrained checkpoints, we construct the Stage-I coarse condition by unrolling the Stage-I sampling for a small number of reverse diffusion steps, consistent with the test-time configuration. Meanwhile, the same sinogram-based DC correction is applied at each Stage-I denoising step, yielding a physically corrected coarse estimate $x_{\text{meta},{\theta1}}$.}

{Given $x_{\text{meta},{\theta1}}$ and the metadata tokens $z_\text{tx}$, Stage-II is then optimized using the EDM v-prediction objective. For computational efficiency, Stage-II is trained without unrolling its sampling trajectory and without applying DC. At each iteration, we sample a diffusion time $t_i$, form the noisy input $x^*_{t_i}$ from the ground-truth target via the forward noising process, and minimize the velocity regression loss:
\begin{equation}
L_{SFT}=\|v_{\theta_2}(x^*_{t_i},t_i,x_{\text{meta},\theta1},z_{\text{tx}})-v_{gt}(x^*_{t_i},t_i)\|_2^2,
\label{eq_new8}
\end{equation}
where $v_{gt}$ denotes the analytical velocity target determined by the diffusion schedule. Gradients are back-propagated through the conditioning path $x_{\text{meta},{\theta1}}$, so that $\theta_2$ is updated directly and $\theta_1$ is updated through its influence on the Stage-II conditioning distribution. Thus, Stage-II is exposed to the actual distribution of Stage-I predictions encountered at inference time, while Stage-I is simultaneously encouraged to generate outputs that are more compatible with Stage-II refinement, thereby allowing residual Stage-I errors to be corrected rather than frozen.}

{In practice, we initialize SFT from the independently trained checkpoints, freeze the text encoder, and use a small learning rate to ensure stable joint optimization.}

\section{Experiments Setup}
\label{sec:experiments}
We evaluate the proposed method on a public multimodal CT dataset (CTRATE) and a real cardiac CT dataset under a range of limited-angle configurations. The data preparation process, including acquisition, preprocessing, metadata extraction, and forward projection simulation, has been detailed in Section~\ref{sec:data_curation}. This section summarizes the competing methods, implementation details, and evaluation metrics.
\subsubsection{Competing Methods}
{To evaluate the effectiveness of our proposed framework, we compare against a comprehensive set of traditional and learning-based baselines, including FBP, ADMM-TV\cite{gao2018combined}, FBPConvNet\cite{jin2017deep}, and recent diffusion models. DOLCE\cite{liu2023dolce} is a state-of-the-art (SOTA) conditional diffusion model for limited-angle CT, which incorporates sinogram and image conditions into a score-based generative model with learnable priors and projection consistency. We reproduce DOLCE with the official code and adapt it to our 2.5D dataset and angular settings for a fair comparison. In contrast, Decomposed Diffusion Sampler (DDS)\cite{chung2023decomposed} is a recent unconditional diffusion model to accelerate large-scale inverse problems. It separates the sampling process from data consistency to flexibly incorporate forward models. We compare against DDS to evaluate the benefit of conditioning on metadata priors. For fairness, all learning-based baselines are retrained and adapted using the same training and testing splits, and evaluated under consistent angular truncation masks and the same measured sinogram. {In addition, whenever forward/backprojection is involved, we use the same CTLIB implementation instantiated with the corresponding scanner geometry}.
}

\subsubsection{Implementation Details}
{All experiments are implemented in PyTorch 2.6.0. Model training is performed on Jetstream2 at Indiana University with a single NVIDIA A100 GPU (40 GB) from the Advanced Cyberinfrastructure Coordination Ecosystem: Services \& Support program. Unless otherwise stated, evaluation and inference are performed on a consumer-grade NVIDIA GeForce RTX 5090 GPU. For the proposed cascaded diffusion model, Stage-I and Stage-II networks are trained independently, each for 400,000 iterations with a batch size of 2, using the Adam optimizer with an initial learning rate of $5.0\times10^{-5}$. Both stages operate on 2.5D slabs. Stage-I is trained with low-resolution slab targets of size $128 \times 128 \times 9$ using a 3D backbone, while Stage-II is trained with high-resolution targets of size $512 \times 512 \times 3$, where the slab depth is represented as channels in the 2D backbone. All learning-based baselines are trained to predict full-resolution outputs ($512 \times 512 \times 3$) following their native formulation.

{After stage-wise pretraining, we perform SFT by unrolling Stage-I sampling for 4 reverse steps with DC to generate the coarse condition and jointly updating both stages for 50,000 iterations with batch size 1 and learning rate $1.0\times10^{-6}$. During SFT, the text encoder is frozen to stabilize optimization.}}
In the DC module, we run a fixed number of ADMM iterations per DC call. We set the number of ADMM iterations per DC call to 10 for Stage-I applied during SFT and inference, and to 30 for Stage-II applied only during inference. The DC hyperparameters are selected empirically and kept fixed across experiments, with $\lambda=1.0$ and $\rho=0.01$ for both stages. At inference, we use 4 sampling steps for Stage-I and 25 steps for Stage-II.

\subsubsection{Evaluation Metrics}
{We evaluate reconstruction quality using both image-based and clinically relevant metrics. For image fidelity, we report five common metrics, including peak signal-to-noise ratio (PSNR), structural similarity index (SSIM), {root-mean-square error in Hounsfield Units (RMSE$_{\text{HU}}$)}, Pearson correlation coefficient (PCC), and normalized mutual information (nMI)\cite{studholme1999overlap}. nMI quantifies the shared information content between the reconstructed image and the reference. In our implementation, nMI is computed with scikit-image, ranging from 1 to 2, with higher values reflecting greater similarity in intensity distributions and structural patterns. 
For the real cardiac CT study, we additionally evaluate clinical utility using coronary artery calcium (CAC) scoring. Specifically, CAC scores are computed from the reconstructed volumes using an automated calcium scoring tool (AI-CAC)\cite{AIoa2400937} and compared against full-view reconstructions to quantify the accuracy of calcium quantification and downstream risk stratification under limited-angle acquisition.

All image-based metrics are reported as mean $\pm$ standard deviation (SD). For the main comparisons, statistics are aggregated across testing cases, while slice-level aggregation is used for ablation studies when appropriate.}

\section{Results}
\label{sec:results}
\subsection{CTRATE Dataset}
\begin{figure*}[!htbp]
\centerline{\includegraphics[width=6.2in]{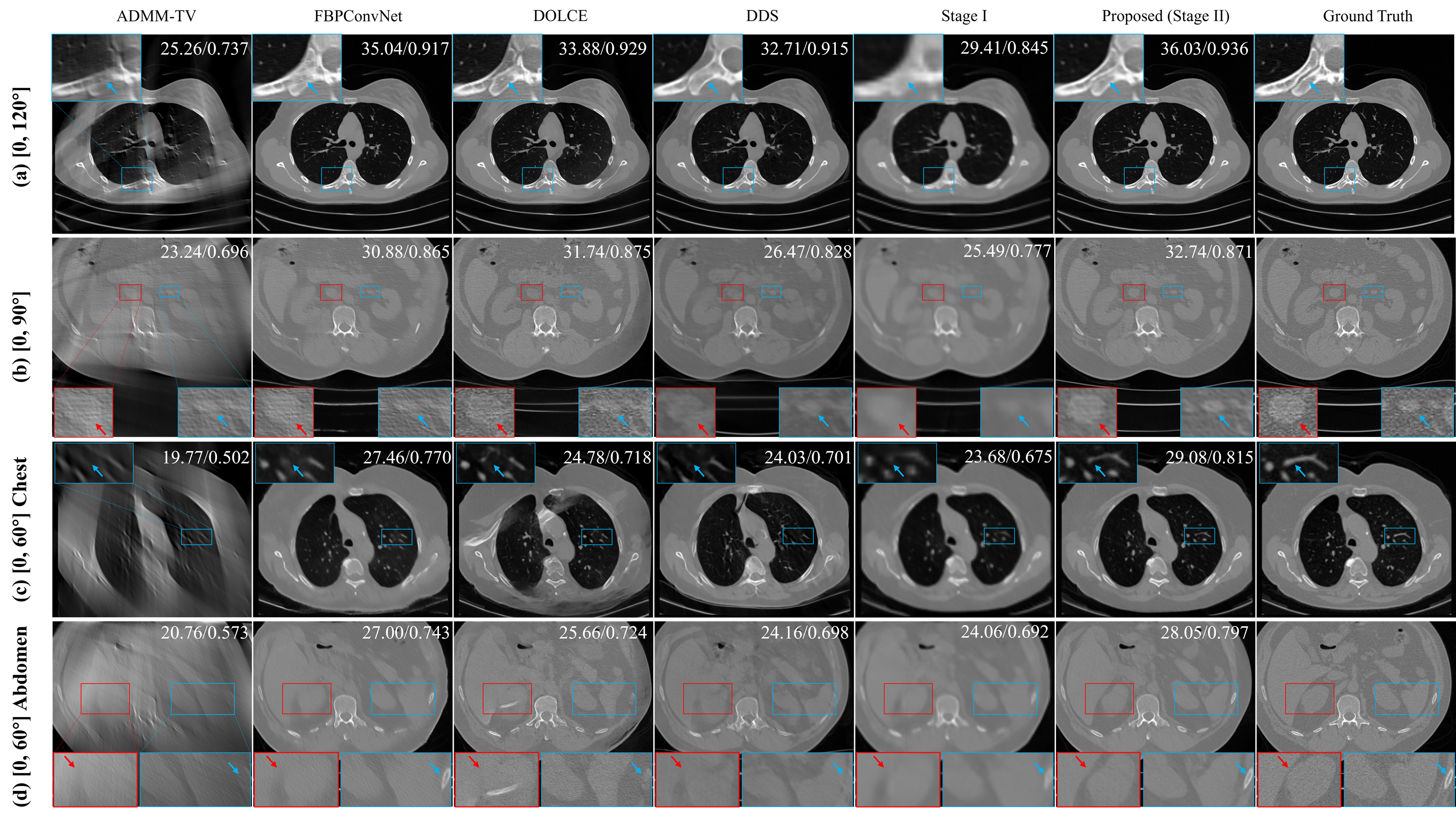}}
\caption{{Qualitative comparison for representative baselines in fan-beam LACT on the CTRATE dataset. (a)–(d) correspond to 120{\degree}, 90{\degree}, 60{\degree} (chest), 60{\degree} (abdomen) cases, respectively. Each subgraph shows reconstructions by ADMM-TV, FBPConvNet, DOLCE, DDS, proposed Stage-I, proposed complete framework, and the full-angle reference. Boxes indicate magnified ROIs with obvious artifacts or hallucinations, and arrows highlight noticeable artifacts or local inconsistencies within the ROIs. All slices are displayed with a $[-1000, 1000]$ Hounsfield unit (HU) display window. PSNR/SSIM are annotated.}}
\label{fig4}
\end{figure*}

Fig. \ref{fig4} presents qualitative comparisons of reconstructions via different methods across 120{\degree}, 90{\degree}, and 60{\degree} LACT settings. Each column images are reconstructions from different methods, with magnified region-of-interests (ROIs)  highlighting anatomical differences.

For 120{\degree} data acquisition, all learning-based methods can recover the major anatomical structures and markedly suppress limited-angle streaks. The CNN-based baseline (FBPConvNet) reconstructs the overall lung appearance well, but exhibits mild over-smoothing and loss of fine cortical details in the magnified bone ROI (blue box in Fig.\ref{fig4}(a)). Diffusion-based approaches further improve local detail recovery while reducing residual artifacts. In particular, the unsupervised DDS effectively suppresses streak-like artifacts, yet may introduce spurious textures that deviate from the full-angle reference. DOLCE provides competitive reconstructions with fewer spurious patterns, but still shows a slight mismatch along the cortical bone boundary. Stage-I offers a reasonable coarse estimate but lacks fine details. By comparison, the proposed complete framework yields the closest match to the references, preserving structural continuity while maintaining effective artifact suppression.

When angle coverage becomes narrow, the advantage of semantic guidance becomes more pronounced. In the 90{\degree} case, the differences are mainly reflected in low-contrast soft-tissue regions (Fig.\ref{fig4}(b)). Supervised methods (FBPConvNet and DOLCE) effectively suppress severe streak artifacts, but tend to over-smooth subtle textures, thereby obscuring fine details in the highlighted ROIs. In contrast, the unsupervised DDS produces sharper edges and more aggressive streak removal, yet is prone to generating hallucinated details. Overall, the proposed method achieves the best balance, yielding soft-tissue appearance and contrast transitions that are most consistent with the reference while keeping residual artifacts minimal. 
When the angle further decreases to 60{\degree}, recovery becomes markedly more challenging for both chest and abdomen. In the chest example (Fig. \ref{fig4}(c)), FBPConvNet exhibits pronounced blurring and loss of subtle structural contrast, DOLCE exhibits prior-driven texture bias (model drift) and residual truncation artifacts, while DDS shows spurious details. In the abdomen example (Fig.~\ref{fig4}(d)), limited-angle streaks and intensity bias are further amplified in relatively homogeneous soft-tissue regions, where the baselines either over-smooth or produce non-uniform local patterns in the ROIs. Across both 60{\degree} cases, the proposed method remains the closest to the full-angle references, although residual discrepancies are still expected under such severe information loss.

Table \ref{tab:quantitative_1} summarizes quantitative results on the CTRATE testing set. As expected, learning-based approaches outperform analytical and iterative baselines. FBPConvNet yields high nMI and PCC from its ability to capture global contrast and coarse structure with large training data. However, as the angular range decreases, its SSIM and PSNR drop and {RMSE$_{\text{HU}}$} increases sharply, indicating reduced robustness in texture and intensity recovery under severe truncation. DDS shows sharper edge recovery, consistent with Fig. \ref{fig4}, but suffers from larger errors in soft-tissue regions, resulting in lower SSIM/PSNR and higher {RMSE$_{\text{HU}}$}. DOLCE remains competitive at moderate truncation but exhibits instability across the CTRATE testing set due to the difficulty of tuning hyperparameters across diverse cases, \emph{e.g.}, $0.840\pm0.172$ for SSIM at 90{\degree}. In contrast, our method consistently ranks first or tied across all metrics and angular settings. Compared to DOLCE (the strongest metadata-free diffusion baseline), it improves SSIM/PSNR by 5.23\%/11.21\% at 90{\degree} and 4.26\%/12.86\% at 60{\degree}, showing the added value of metadata integration via a two-stage transformer-based UNet for LACT imaging.

\begin{figure*}[!htb]
\centerline{\includegraphics[width=6.2in]{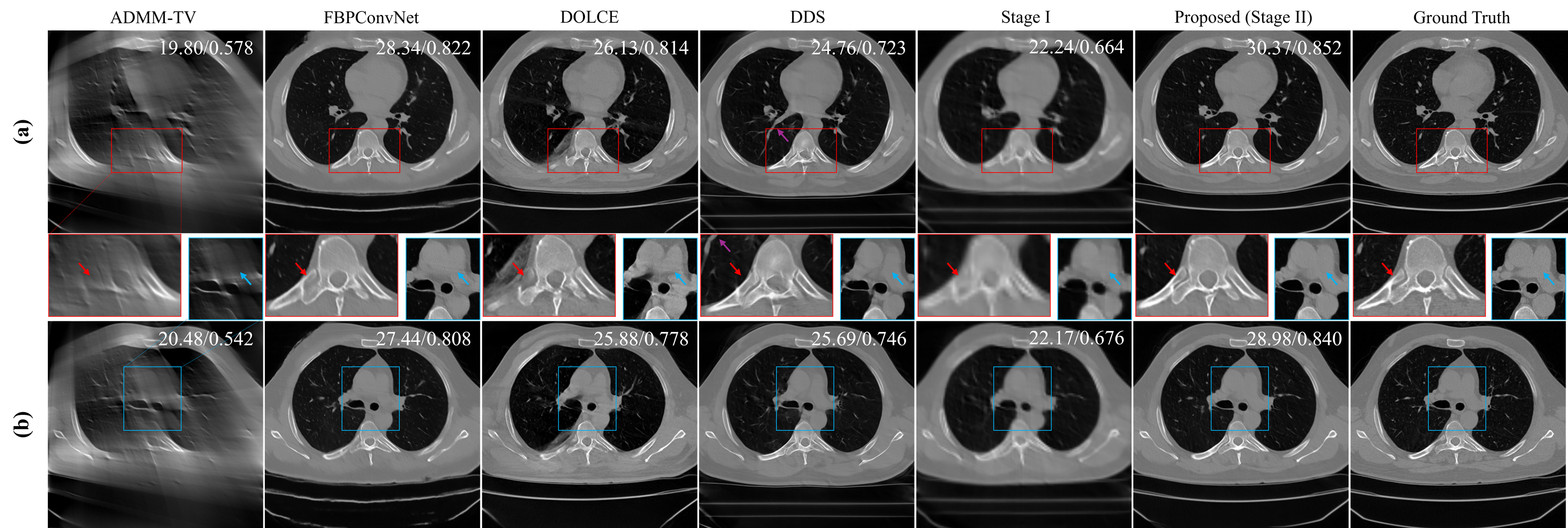}}
\caption{{Qualitative comparison with representative baselines for fan-beam LACT under an unseen 75{\degree} scan on the CTRATE dataset. (a) and (b) show two representative slices. In (a), the red-box indicates a magnified high-contrast bony ROI; in (b), the blue-box indicates a magnified a mediastinal soft-tissue ROI containing major vessels and airway structures. Each subfigure includes reconstructions from ADMM-TV, FBPConvNet, DOLCE, DDS, proposed Stage-I, proposed complete framework (Stage-II), and the full-angle reference. Arrows highlight noticeable differences in the ROIs. The display window is $[-1000, 1000]$ HU for the full slices and the bone ROIs, and $[-700, 700]$ HU for the zoomed-in mediastinal ROI. PSNR/SSIM are annotated.}}
\label{fig5}
\vspace{-3mm}
\end{figure*}

\begin{table}[t]
\centering
\caption{Quantitative comparison of reconstruction methods on CTRATE datasets under limited-angle scenarios seen during training.}
\resizebox{\columnwidth}{!}{
\begin{tabular}{l ccccc}
\toprule
\textbf{Method} & SSIM & PSNR & nMI & PCC & {$\mathrm{RMSE}_{\mathrm{HU}}$} \\
\midrule
\multicolumn{6}{c}{\textbf{120{\degree}}}\\
\midrule
FBP~           & $0.509\pm0.042$ & $17.274\pm1.829$ & $1.162\pm0.010$ & $0.862\pm0.026$ & $219.0\pm23.8$ \\
ADMM-TV~       & $0.704\pm0.027$ & $26.609\pm1.136$ & $1.199\pm0.012$ & $0.931\pm0.061$ & $71.9\pm8.9$ \\
FBPConvNet~    & $\underline{\emph{0.905}\pm\emph{0.022}}$ & $34.642\pm1.583$ & $1.325\pm0.027$ & $0.983\pm0.024$ & $\underline{\emph{29.1}\pm\emph{2.6}}$ \\
DOLCE~         & $\textbf{0.928}\pm\textbf{0.025}$ & $\underline{\emph{35.550}\pm\emph{3.930}}$ & $\underline{\emph{1.330}\pm\emph{0.072}}$ & $\underline{\emph{0.989}\pm\emph{0.004}}$ & $31.3\pm9.2$ \\
DDS~           & $\underline{\emph{0.905}\pm\emph{0.041}}$ & $31.622\pm3.261$ & $1.253\pm0.055$ & $0.942\pm0.004$ & $48.7\pm13.6$ \\
\textbf{Proposed}~  & $\textbf{0.928}\pm\textbf{0.018}$ & $\textbf{36.437}\pm\textbf{1.520}$ & $\textbf{1.352}\pm\textbf{0.026}$ & $\textbf{0.994}\pm\textbf{0.002}$ & $\textbf{25.1}\pm\textbf{3.6}$ \\
\midrule
\multicolumn{6}{c}{\textbf{90{\degree}}}\\
\midrule
FBP~           & $0.415\pm0.047$ & $15.968\pm1.822$ & $1.127\pm0.007$ & $0.681\pm0.041$ & $254.6\pm27.5$ \\
ADMM-TV~       & $0.598\pm0.024$ & $23.572\pm1.045$ & $1.177\pm0.011$ & $0.876\pm0.063$ & $101.8\pm11.7$ \\
FBPConvNet~    & $\underline{\emph{0.840}\pm\emph{0.030}}$ & $30.011\pm1.650$ & $1.281\pm0.023$ & $\underline{\emph{0.965}\pm\emph{0.027}}$ & $\underline{\emph{49.8}\pm\emph{8.3}}$\\
DOLCE~         & $\underline{\emph{0.840}\pm\emph{0.172}}$ & $\underline{\emph{30.245}\pm\emph{4.507}}$ & $\underline{\emph{1.285}\pm\emph{0.069}}$ & $0.961\pm0.066$ & $55.3\pm39.2$ \\
DDS~           & $0.794\pm0.057$ & $26.766\pm3.627$ & $1.225\pm0.033$ & $0.914\pm0.012$ & $82.9\pm20.5$ \\
\textbf{Proposed}~  & $\textbf{0.884}\pm\textbf{0.027}$ & $\textbf{33.634}\pm\textbf{1.802}$ & $\textbf{1.314}\pm\textbf{0.024}$ & $\textbf{0.988}\pm\textbf{0.003}$ & $\textbf{34.8}\pm\textbf{5.6}$ \\
\midrule
\multicolumn{6}{c}{\textbf{60{\degree}}}\\
\midrule
FBP~           & $0.350\pm0.059$ & $14.981\pm1.813$ & $1.104\pm0.005$ & $0.430\pm0.068$ & $285.4\pm30.8$\\
ADMM-TV~       & $0.505\pm0.031$ & $21.203\pm1.100$ & $1.163\pm0.012$ & $0.792\pm0.073$ & $133.9\pm16.5$\\
FBPConvNet~    & $0.773\pm0.043$ & $\underline{\emph{27.266}\pm\emph{1.608}}$ & $\underline{\emph{1.254}\pm\emph{0.021}}$ & $0.940\pm0.037$ & $\underline{\emph{68.4}\pm\emph{10.2}}$\\
DOLCE~         & $\underline{\emph{0.774}\pm\emph{0.152}}$ & $26.037\pm5.129$ & $1.240\pm0.035$ & $\underline{\emph{0.942}\pm\emph{0.023}}$ & $82.4\pm17.7$\\
DDS~           & $0.707\pm0.059$ & $25.157\pm2.954$ & $1.192\pm0.026$ & $0.826\pm0.038$ & $100.1\pm16.8$\\
\textbf{Proposed}~  & $\textbf{0.807}\pm\textbf{0.035}$ & $\textbf{29.386}\pm\textbf{1.714}$ & $\textbf{1.281}\pm\textbf{0.020}$ & $\textbf{0.970}\pm\textbf{0.008}$ & $\textbf{56.1}\pm\textbf{8.3}$\\
\bottomrule
\end{tabular}
}
\label{tab:quantitative_1}
\vspace{-2mm}
\end{table}
To evaluate the generalizability of the proposed method, we conduct two out-of-distribution experiments on CTRATE: (i) fan-beam CT projection with a 75{\degree} angular range that is not included in training, and (ii) 45{\degree} parallel-beam projections with both unseen angle and geometry. Results of the first experiment are presented in Fig.\ref{fig5}.
In the high-contrast bony region (Fig.\ref{fig5} (a)), FBPConvNet reconstructs the gross structure of the thoracic vertebrae but blurs fine cortical structures, failing to clearly separate the interspaces between the vertebrae and ribs (red arrows), and it degrades in severely truncated areas.
Under this unseen acquisition setting, DOLCE exhibits ground-glass-like artifacts near the ribs, indicating reduced robustness to domain shift. DDS improves edge sharpness and bone contrast but generates a spurious vessel-like tubular structure in the lung region, as highlighted by purple arrow in Fig.\ref{fig5} (a). Our method provides sharp vertebral reconstructions and clear separation between vertebrae and ribs with minimal hallucinations. 
For the mediastinal ROI containing airway lumen and adjacent great-vessel structures (Fig.\ref{fig5}(b)), supervised baselines suppress streak artifacts but tend to smooth low-contrast boundaries and subtle contrast transitions. DDS yields higher apparent sharpness, yet shows local structural deviations from the full-angle reference in the magnified view. The proposed method better preserves the airway/vessel delineation and soft-tissue contrast, achieving the closest visual agreement with the reference.

Quantitative results in Table~\ref{tab:quantitative_2} further confirm the qualitative comparisons in Fig.~\ref{fig5} under the unseen 75{\degree} fan-beam setting. The proposed method achieves the best performance across all metrics. Among the baselines, DOLCE is the strongest competitor in terms of SSIM and nMI, whereas FBPConvNet yields the second-best PSNR and RMSE$_{\text{HU}}$; the two methods are comparable on PCC. Relative to the second-best method for each metric, the proposed method improves SSIM by 2.21\%, reduces RMSE$_{\text{HU}}$ by 25.39\%, and increases PSNR, nMI, and PCC by 10.81\%, 1.42\%, and 2.73\%, respectively. Although 75{\degree} is not seen during training, it lies between the two fan-beam angles (60{\degree} and 90{\degree}) used for training, corresponding to a mild distribution shift. Nevertheless, our method remains the most robust and accurate.

\begin{figure*}[!htbp]
\centerline{\includegraphics[width=6.2in]{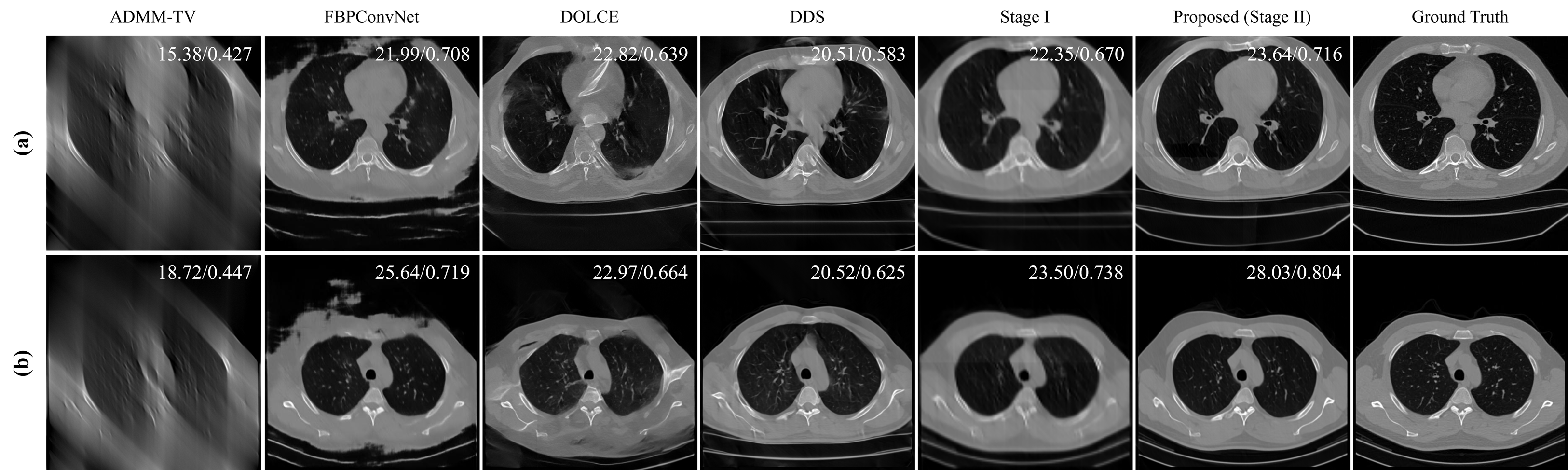}}
\caption{{Qualitative comparison of the SOTA methods for parallel-beam LACT with 45{\degree} scan. (a) and (b) show two representative slices under this setting. All images are displayed with the same window $[-1000, 1000]$ HU. PSNR/SSIM are annotated.}
}
\label{fig6}
\end{figure*}

\begin{table}[t]
\centering
\caption{Quantitative comparison on the CTRATE under unseen LACT acquisition settings.}
\resizebox{0.41\textwidth}{!}{
\begin{tabular}{l ccccc}
\toprule
\textbf{Method} & SSIM & PSNR & nMI & PCC & {RMSE$_{\text{HU}}$}\\
\midrule
\multicolumn{6}{c}{\textbf{75$^\circ$ (fan-beam)}} \\
\midrule
FBP            & $0.368\pm0.036$ & $15.135\pm1.603$ & $1.110\pm0.006$ & $0.532\pm0.061$ & $279.6\pm18.9$ \\
ADMM-TV        & $0.532\pm0.029$ & $21.842\pm1.099$ & $1.175\pm0.011$ & $0.837\pm0.066$ & $117.2\pm14.4$ \\
FBPConvNet     & $0.797\pm0.034$ & $\underline{\emph{27.823}\pm\emph{1.437}}$ & $1.267\pm0.021$ & $\underline{\emph{0.952}\pm\emph{0.019}}$ & $\underline{\emph{63.8}\pm\emph{9.2}}$ \\
DOLCE          & $\underline{\emph{0.816}\pm\emph{0.044}}$ & $27.456\pm3.802$ & $\underline{\emph{1.271}\pm\emph{0.035}}$ & $\underline{\emph{0.952}\pm\emph{0.018}}$ & $68.2\pm15.3$ \\
DDS            & $0.750\pm0.065$ & $24.683\pm2.994$ & $1.239\pm0.026$ & $0.938\pm0.017$ & $90.2\pm22.5$ \\
\textbf{Proposed} & $\textbf{0.834}\pm\textbf{0.028}$ & $\textbf{30.832}\pm\textbf{1.841}$ & $\textbf{1.289}\pm\textbf{0.017}$ & $\textbf{0.978}\pm\textbf{0.005}$ & $\textbf{47.6}\pm\textbf{7.7}$ \\
\midrule
\multicolumn{6}{c}{\textbf{45$^\circ$ (parallel-beam)}} \\
\midrule
FBP            & $0.324\pm0.065$ & $14.586\pm1.826$ & $1.099\pm0.006$ & $0.375\pm0.076$ & $295.6\pm32.0$ \\
ADMM-TV        & $0.434\pm0.025$ & $17.579\pm1.081$ & $1.151\pm0.013$ & $0.770\pm0.082$ & $140.3\pm17.8$ \\
FBPConvNet     & $\underline{\emph{0.698}\pm\emph{0.045}}$ & $\underline{\emph{24.055}\pm\emph{1.538}}$ & $1.206\pm0.016$ & $\underline{\emph{0.888}\pm\emph{0.040}}$ & $\underline{\emph{98.3}\pm\emph{12.2}}$ \\
DOLCE          & $0.669\pm0.078$ & $23.448\pm5.923$ & $\underline{\emph{1.210}\pm\emph{0.032}}$ & $0.870\pm0.078$ & $102.7\pm37.3$ \\
DDS            & $0.651\pm0.035$ & $22.807\pm2.660$ & $1.161\pm0.018$ & $0.794\pm0.127$ & $134.6\pm12.6$ \\
\textbf{Proposed} & $\textbf{0.744}\pm\textbf{0.039}$ & $\textbf{26.144}\pm\textbf{1.978}$ & $\textbf{1.225}\pm\textbf{0.017}$ & $\textbf{0.935}\pm\textbf{0.018}$ & $\textbf{80.8}\pm\textbf{12.2}$ \\
\bottomrule
\end{tabular}}
\label{tab:quantitative_2}
\vspace{-4mm}
\end{table}
The second experiment is more challenging, involving both an unseen angular range (45{\degree}) and a new acquisition geometry (parallel-beam). As shown in Fig.\ref{fig6}, all baselines exhibit noticeable degradation under this substantial distribution shift. FBPConvNet introduces pronounced artifacts in the data-missing regions, while DOLCE reconstructs object sizes roughly but fails to suppress boundary artifacts and exhibits banding and brightness fluctuations across soft-tissue regions. DDS yields perceptually cleaner results than supervised methods but still suffers from shape distortion and inaccurate boundaries, indicating its limited capability in preserving accurate spatial geometry under substantial distribution shifts. In contrast, our method achieves the highest visual fidelity, with structures closely matching the references in both perceptual quality and anatomy. These observations are confirmed by the quantitative results in Table \ref{tab:quantitative_2}, where our method outperforms the second-best method by 6.59\%, 17.80\%, 8.68\%, 1.24\%, and 5.29\% on SSIM, {RMSE$_{\text{HU}}$}, PSNR, nMI, and PCC, respectively.
Furthermore, the 45{\degree} parallel-beam setting remains extremely challenging due to both the unseen angular range and the unseen acquisition geometry. Although our method achieves the best quantitative performance in Table~\ref{tab:quantitative_2}, a substantial gap to the full-angle reference still persists, as evidenced by the noticeable structural deviations and residual artifacts in Fig.~\ref{fig6}(a). This indicates that the missing angular information under such an extreme configuration cannot be fully compensated even with strong learned priors. Therefore, we treat the 45{\degree} parallel-beam experiment as a strict robustness benchmark for performance evaluation, rather than a practically deployable clinical acquisition protocol.

Beyond comparisons with the reconstruction methods, Figs. \ref{fig4}--\ref{fig6} include intermediate results from Stage-I of our framework. Although these initial reconstructions appear relatively blurry, they already exhibit reduced limited-angle artifacts, demonstrating the effectiveness of metadata-driven priors in guiding early-stage reconstructions.

\subsection{Real Cardiac Dataset}
To further evaluate generalizability, we test the proposed approach on real cardiac CT dataset acquired on a GE scanner. Due to differences in scanner hardware, acquisition protocols, and patient populations, the GE dataset introduces a noticeable domain shift relative to CTRATE. In addition, linear normalization is applied to the raw sinograms to align their dynamic range with the training data.

\begin{figure}[!htb]
\vspace{-1mm}
\centerline{\includegraphics[width=3.1in]{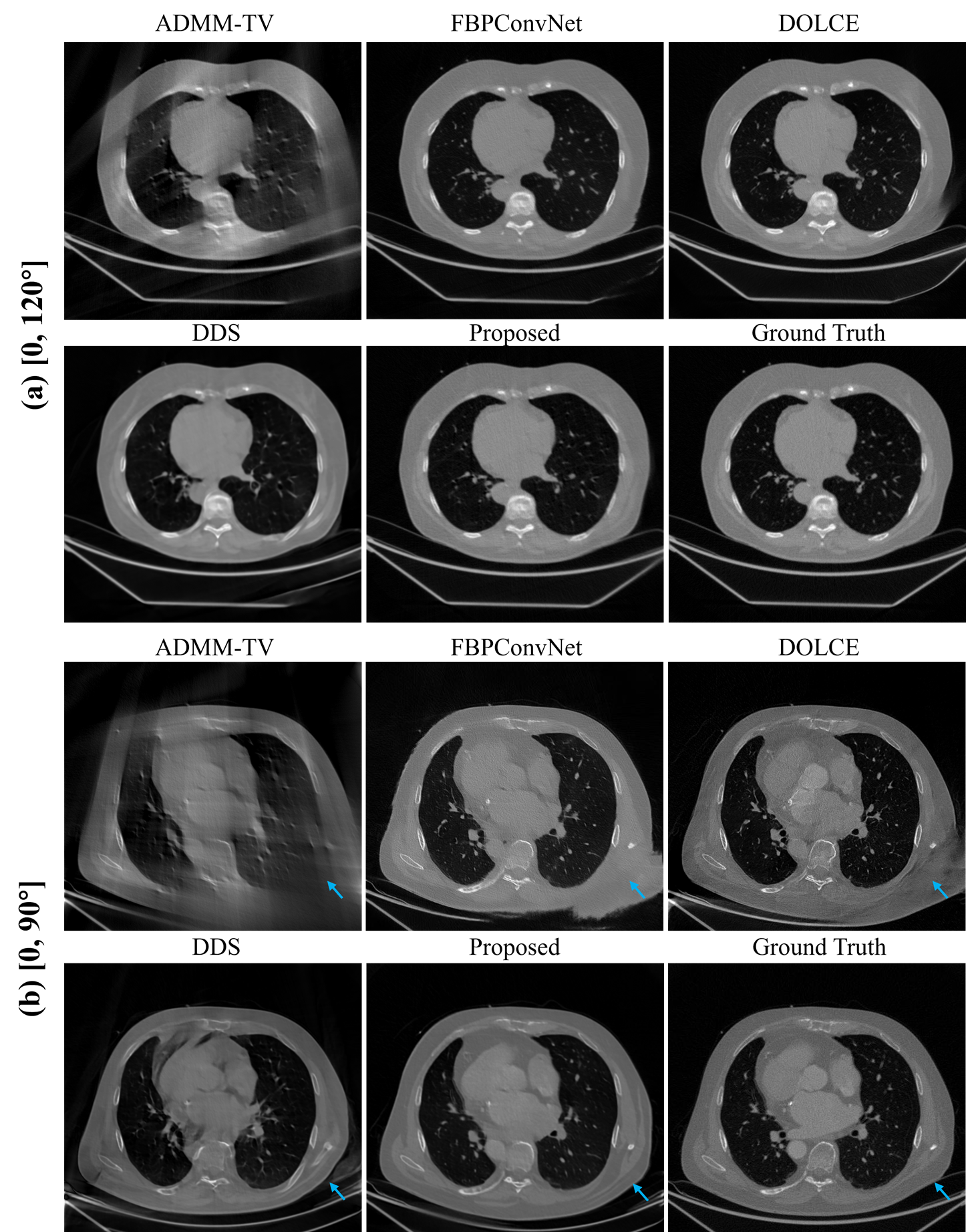}}
\caption{{Clinical LACT reconstructions on GE cardiac CT under two angular coverages: (a) 120{\degree}; (b) 90{\degree}. Reconstructions are shown for ADMM-TV, FBPConvNet, DOLCE, DDS, Proposed (complete framework), and the full-view reference. Arrows highlight residual limited-angle artifacts notably at 90{\degree}.} All images are displayed with a $[-1000, 1000]$ HU window.
}
\vspace{-4mm}
\label{fig7}
\end{figure}

Fig.~\ref{fig7} shows representative reconstructions under two angular coverages (120{\degree} and 90{\degree}). As expected, the traditional iterative methods, such as ADMM-TV, are relatively robust to data distribution shifts, producing results with consistent visual characteristics. After sinogram normalization, FBPConvNet remains a strong supervised baseline on this GE cohort, but still exhibits mild shading and boundary inconsistencies in severely truncated regions. Nevertheless, since FBPConvNet preserves global structural integrity to some extent, its metrics remain relatively high. 
DOLCE, as a supervised model, benefits from an explicit data consistency term that partially mitigates global intensity drift. However, it leaves more residual limited-angle artifacts than the CTRATE dataset (\emph{e.g.}, Fig.\ref{fig4}(b)). DDS suppresses some streak-like artifacts, but shows reduced structural consistency under the scanner-domain shift. By incorporating metadata-guided priors with data consistency, the proposed method reconstructs images that are most consistent with the full-angle reference across both angular coverages, which is also reflected by the best quantitative performance in Table~\ref{tab:quantitative_clinic}.

\begin{table}[t]
\centering
\caption{Quantitative comparison of reconstruction methods on clinical GE LACT dataset.}
\resizebox{0.41\textwidth}{!}{
\begin{tabular}{l ccccc}
\toprule
\textbf{Method} & SSIM & PSNR & nMI & PCC & {RMSE$_{\text{HU}}$} \\
\midrule
\multicolumn{6}{c}{\textbf{120$^\circ$}} \\
\midrule
FBP            & $0.552\pm0.042$ & $18.547\pm0.923$ & $1.096\pm0.004$ & $0.880\pm0.012$ & $178.1\pm18.6$ \\
ADMM-TV        & $0.741\pm0.019$ & $26.763\pm0.732$ & $1.156\pm0.010$ & $0.919\pm0.009$ & $70.2\pm5.9$ \\
FBPConvNet     & $0.915\pm0.007$ & $\underline{\emph{32.876}\pm\emph{0.442}}$ & $\underline{\emph{1.278}\pm0.007}$ & $\underline{\emph{0.988}\pm0.001}$ & $\underline{\emph{34.1}\pm\emph{1.8}}$ \\
DOLCE          & $\underline{\emph{0.917}\pm\emph{0.028}}$ & $31.923\pm1.431$ & $1.139\pm0.012$ & $0.931\pm0.061$ & $38.6\pm8.2$ \\
DDS            & $0.862\pm0.016$ & $30.096\pm0.738$ & $1.228\pm0.008$ & $0.976\pm0.005$ & $47.1\pm4.1$ \\
\textbf{Proposed} & $\textbf{0.944}\pm\textbf{0.006}$ & $\textbf{36.082}\pm0.980$ & $\textbf{1.320}\pm0.012$ & $\textbf{0.994}\pm0.002$ & $\textbf{24.4}\pm2.8$ \\
\midrule
\multicolumn{6}{c}{\textbf{90$^\circ$}} \\
\midrule
FBP            & $0.468\pm0.040$ & $17.374\pm0.885$ & $1.0631\pm0.002$ & $0.749\pm0.017$ & $203.8\pm20.5$ \\
ADMM-TV        & $0.655\pm0.020$ & $23.002\pm0.602$ & $1.113\pm0.007$ & $0.884\pm0.020$ & $107.4\pm7.7$ \\
FBPConvNet     & $\underline{\emph{0.855}\pm\emph{0.011}}$ & $\underline{\emph{28.108}\pm\emph{0.673}}$ & $\underline{\emph{1.219}\pm\emph{0.006}}$ & $\underline{\emph{0.962}\pm\emph{0.005}}$ & $\underline{\emph{59.2}\pm\emph{4.1}}$ \\
DOLCE          & $0.826\pm0.013$ & $26.456\pm0.498$ & $1.189\pm0.005$ & $0.946\pm0.006$ & $71.5\pm4.0$ \\
DDS            & $0.761\pm0.023$ & $24.668\pm0.688$ & $1.159\pm0.008$ & $0.914\pm0.017$ & $87.9\pm7.0$ \\
\textbf{Proposed} & $\textbf{0.885}\pm\textbf{0.014}$ & $\textbf{30.033}\pm\textbf{0.861}$ & $\textbf{1.238}\pm\textbf{0.010}$ & $\textbf{0.976}\pm\textbf{0.003}$ & $\textbf{48.2}\pm\textbf{4.9}$ \\
\bottomrule
\end{tabular}}
\label{tab:quantitative_clinic}
\vspace{-3mm}
\end{table}

\begin{figure}[!htb]
\vspace{-1mm}
\centerline{\includegraphics[width=3.4in]{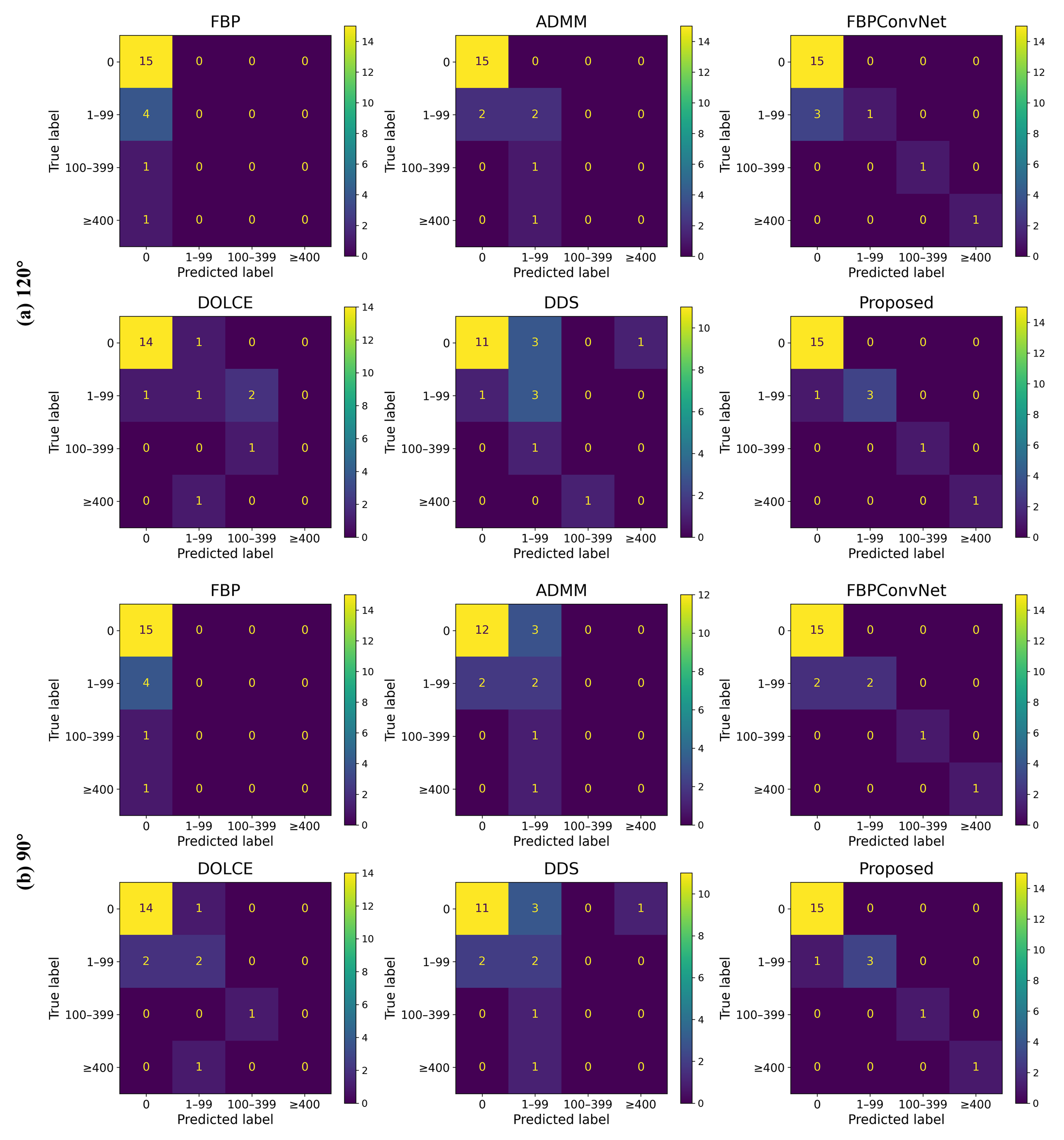}}
\caption{{CAC risk-stratification consistency on the clinical non-contrast cohort (n=21). Confusion matrices (0, 1--99, 100--399, $\geq$400) compare CAC categories computed by AI-CAC on reconstructions from different methods against the reference CAC categories obtained from the full-angle reconstructions.}
}
\vspace{-4mm}
\label{fig8}
\end{figure}

{To demonstrate the potential clinical impacts, we further evaluate CAD-relevant calcified-plaque burden on the non-contrast CAC cohort using AI-CAC\cite{AIoa2400937}. Because the manual CAC annotations are unavailable, we adopt the CAC risk categories produced by AI-CAC on the full-angle reconstruction as a pseudo-references. We then run AI-CAC on the reconstructions from different methods under limited-angle measurements and report 4-class CAC risk-category confusion matrices (0, 1--99, 100--399, and $\geq$400)\cite{Pletcher2004}. The reference cohort includes 15, 4, 1, and 1 cases in the four categories, respectively. As shown in Fig. \ref{fig8}, FBP largely collapses non-zero CAC cases into the zero category, indicating severe underestimation under limited-angle acquisition. ADMM-TV and learning-based baselines alleviate this issue to different degrees but still exhibit both up- and down-grading errors. In contrast, the proposed method achieves the most consistent risk-category assignments with the pseudo-references, correctly preserving the two high-risk cases (100--399 and $\geq$400) and yielding only minimal category shifts for the mild-risk group. These results suggest preservation of CAC risk stratification despite the information loss in limited-angle reconstruction.}

{Detailed metadata including acquisition parameters and de-identified clinical context for the representative cases in Figs. \ref{fig4}--\ref{fig7} are provided in the Supplementary Material\footnote{Supplementary materials are available in the supporting documents/multimedia tab.}
.}

\subsection{Ablation Study}
\subsubsection{{Impacts of Clinical Metadata}}
We study how clinical metadata influences LACT reconstruction quality on CTRATE by ablating three metadata categories: physical scanning parameters (tube current, exposure time, bed position and acquisition angles), demographics (patient gender and age), and diagnostic information (predicted diseases and physician impressions), denoted as \emph{Phy}, \emph{Demo}, and \emph{Diag}, respectively. {We evaluate five variants: (i) no-metadata baseline where the T5X cross-attention receives a null token, (ii) three single-category ablations that remove one metadata group at a time, and (iii) the full-metadata model. All variants share the same network backbone, ADMM/data-consistency configuration, sampling steps, and training protocol to isolate the effects of metadata.} Quantitative results are summarized in Table \ref{tab:ablation} and reported as slice-level statistics.

{Overall, incorporating metadata yields consistent improvements across truncation levels. The full-metadata model achieves the best mean PSNR/SSIM under all truncation angles. 
More importantly, paired two-sided t-tests within each angle group confirm that metadata conditioning achieves statistically significant improvements in both PSNR and SSIM over the no-metadata variant (p $<$ 0.05). At 60{\degree} truncated-scan condition, metadata improves PSNR by 0.411 dB (95\% CI [0.239, 0.581]) and improves SSIM by 0.015 (95\% CI [0.012, 0.018]). At 90{\degree}, PSNR increases by 0.518 dB (95\% CI [0.382, 0.655]) and SSIM increases by 0.018 (95\% CI [0.016, 0.019]). At 120{\degree}, PSNR increases by 0.318 dB (95\% CI [0.190, 0.448]) and SSIM increases by 0.012 (95\% CI [0.011, 0.013]). These results show that the improvements are statistically reliable rather than within random variation.}

\begin{table*}[t]
\centering
\caption{Ablation study results on metadata type, network design, frequency of data consistency, and SFT.}
\setlength{\tabcolsep}{6pt}
\renewcommand{\arraystretch}{1.15}
\setlength{\tabcolsep}{4pt}
\begin{tabu}{ccc cc cc cc}
\toprule
\multicolumn{3}{c}{\textbf{Metadata Type}} 
& \multicolumn{2}{c}{\textbf{120$^\circ$}} 
& \multicolumn{2}{c}{\textbf{90$^\circ$}} 
& \multicolumn{2}{c}{\textbf{60$^\circ$}} \\
\cmidrule(lr){1-3} \cmidrule(lr){4-5} \cmidrule(lr){6-7} \cmidrule(lr){8-9}
Demo & Diag & Phy & SSIM & PSNR & SSIM & PSNR & SSIM & PSNR \\
\midrule
\ding{55} & \ding{55} & \ding{55} & $0.923\!\pm\!0.029$ & $36.753\!\pm\!3.741$ & $0.870\!\pm\!0.036$ & $32.960\!\pm\!4.109$ & $0.785\!\pm\!0.054$ & $29.444\!\pm\!4.464$ \\
\ding{55} & \checkmark & \checkmark & $0.930\!\pm\!0.030$ & $36.984\!\pm\!3.732$ & $0.883\!\pm\!0.037$ & $33.234\!\pm\!4.099$ & $0.791\!\pm\!0.053$ & $29.492\!\pm\!4.497$ \\
\checkmark & \ding{55} & \checkmark & $0.927\!\pm\!0.030$ & $36.904\!\pm\!3.726$ & $0.880\!\pm\!0.037$ & $33.309\!\pm\!4.111$ & $0.794\!\pm\!0.053$ & $29.652\!\pm\!4.444$ \\
\checkmark & \checkmark & \ding{55} & $0.924\!\pm\!0.030$ & $36.970\!\pm\!3.727$ & $0.877\!\pm\!0.036$ & $33.158\!\pm\!4.084$ &  $0.788\!\pm\!0.053$ & $29.567\!\pm\!4.468$ \\
\checkmark & \checkmark & \checkmark & $0.935\!\pm\!0.029$ & $37.072\!\pm\!3.739$ & $0.888\!\pm\!0.036$ & $33.479\!\pm\!4.081$ & $0.801\!\pm\!0.053$ & $29.854\!\pm\!4.491$ \\
\midrule
& \multicolumn{1}{c}{\multirow{3}{*}{\textbf{Structure}}} & 
& \multicolumn{2}{c}{\textbf{120$^\circ$}} 
& \multicolumn{2}{c}{\textbf{90$^\circ$}} 
& \multicolumn{2}{c}{\textbf{60$^\circ$}} \\
\cmidrule(lr){4-5} \cmidrule(lr){6-7} \cmidrule(lr){8-9}
 & & & SSIM & PSNR & SSIM & PSNR & SSIM & PSNR \\
\midrule
& \makecell[c]{Single Stage (25 steps)} & & $0.798\!\pm\!0.071$ & $26.293\!\pm\!4.652$ & $0.687\!\pm\!0.077$ & $21.406\!\pm\!5.298$ & $0.600\!\pm\!0.102$ & $21.928\!\pm\!5.797$ \\
& \makecell[c]{Single Stage (100 steps)} & & $0.811\!\pm\!0.060$ & $27.268\!\pm\!4.518$ & $0.733\!\pm\!0.069$ & $22.815\!\pm\!6.230$ & $0.703\!\pm\!0.108$ & $22.866\!\pm\!7.432$\\
& \makecell[c]{Cascaded (Proposed)} & & $0.935\!\pm\!0.029$ & $37.072\!\pm\!3.739$ & $0.888\!\pm\!0.036$ & $33.479\!\pm\!4.081$ & $0.801\!\pm\!0.053$ & $29.854\!\pm\!4.491$\\
\midrule
& \multicolumn{1}{c}{\multirow{3}{*}{\textbf{Frequency of DC}}} &
& \multicolumn{2}{c}{\textbf{120$^\circ$}} 
& \multicolumn{2}{c}{\textbf{90$^\circ$}} 
& \multicolumn{2}{c}{\textbf{60$^\circ$}} \\
\cmidrule(lr){4-5} \cmidrule(lr){6-7} \cmidrule(lr){8-9}
 & & & SSIM & PSNR & SSIM & PSNR & SSIM & PSNR \\
\midrule
& \makecell[c]{w/o DC} & & $0.750\!\pm\!0.086$ & $31.632\!\pm\!3.494$ & $0.716\!\pm\!0.084$ & $30.032\!\pm\!3.591$ & $0.698\!\pm\!0.084$ & $27.993\!\pm\!4.342$ \\
& \makecell[c]{E1 Sampling Steps} & & $0.932\!\pm\!0.029$ & $37.319\!\pm\!3.685$ & $0.891\!\pm\!0.043$ & $34.450\!\pm\!4.353$ & $0.795\!\pm\!0.053$ & $29.457\!\pm\!4.351$\\
& \makecell[c]{E5 Sampling Steps} & & $0.935\!\pm\!0.029$ & $37.072\!\pm\!3.739$ & $0.888\!\pm\!0.036$ & $33.479\!\pm\!4.081$ & $0.801\!\pm\!0.053$ & $29.854\!\pm\!4.491$\\
& \makecell[c]{E10 Sampling Steps} & & $0.924\!\pm\!0.029$ & $36.156\!\pm\!3.789$ & $0.875\!\pm\!0.040$ & $33.248\!\pm\!4.284$ & $0.788\!\pm\!0.058$ & $29.117\!\pm\!4.683$\\
& \makecell[c]{Last Sampling Step} & & $0.905\!\pm\!0.033$ & $35.202\!\pm\!3.912$ & $0.842\!\pm\!0.048$ & $31.742\!\pm\!4.612$ & $0.754\!\pm\!0.062$ & $28.531\!\pm\!4.987$\\
& \makecell[c]{DC Only} & & $0.891\!\pm\!0.018$ & $34.216\!\pm\!1.665$ & $0.810\!\pm\!0.026$ & $30.961\!\pm\!1.537$ & $0.688\!\pm\!0.041$ & $27.320\!\pm\!2.082$\\
\midrule
& \multicolumn{1}{c}{\multirow{3}{*}{\textbf{SFT}}} &
& \multicolumn{2}{c}{\textbf{120$^\circ$}} 
& \multicolumn{2}{c}{\textbf{90$^\circ$}} 
& \multicolumn{2}{c}{\textbf{60$^\circ$}} \\
\cmidrule(lr){4-5} \cmidrule(lr){6-7} \cmidrule(lr){8-9}
 & & & SSIM & PSNR & SSIM & PSNR & SSIM & PSNR \\
\midrule
& \makecell[c]{w/o SFT} & & $0.916\!\pm\!0.028$ & $34.140\!\pm\!3.978$ & $0.817\!\pm\!0.043$ & $30.176\!\pm\!4.561$ & $0.688\!\pm\!0.067$ & $25.692\!\pm\!4.406$ \\
& \makecell[c]{w/ SFT (Proposed)} & & $0.935\!\pm\!0.029$ & $37.072\!\pm\!3.739$ & $0.888\!\pm\!0.036$ & $33.479\!\pm\!4.081$ & $0.801\!\pm\!0.053$ & $29.854\!\pm\!4.491$\\
\bottomrule
\end{tabu}
\label{tab:ablation}
\vspace{-3.0mm}
\end{table*}

When ablating individual categories, removing any single metadata type consistently degrades performance relative to the full-metadata setting, indicating that different metadata sources provide complementary priors. {Among them, \emph{Phy} exhibits the most consistent impact on structural similarity across truncation levels: SSIM drops from 0.801 to 0.788 at 60{\degree}, from 0.888 to 0.877 at 90{\degree}, and from 0.935 to 0.924 at 120{\degree}, with concurrent PSNR decrease, \emph{e.g.}, from 29.854 to 29.567 at 60 {\degree} and 33.479 to 33.158 at 90 {\degree}. These results suggest that scanner-related priors help the model adapt to acquisition-dependent artifact and noise characteristics, and stabilize the refinement.}

{Removing \emph{Demo} or \emph{Diag} also causes systematic performance drops, although the relative impact depends on truncation severity. For example, under the most ill-posed 60{\degree} condition, removing \emph{Demo} leads to the largest PSNR reduction ($\Delta$PSNR=-0.362 dB) together with a noticeable SSIM decrease ($\Delta$SSIM=-0.010), implying that patient-level priors provide useful global constraints when projections are highly incomplete. By contrast, removing \emph{Diag} yields modest yet consistent degradations across angles, suggesting that diagnostic priors contribute additional information to refine anatomy-consistent details.}

\begin{figure}[!htbp]
\centerline{\includegraphics[width=3.0in]{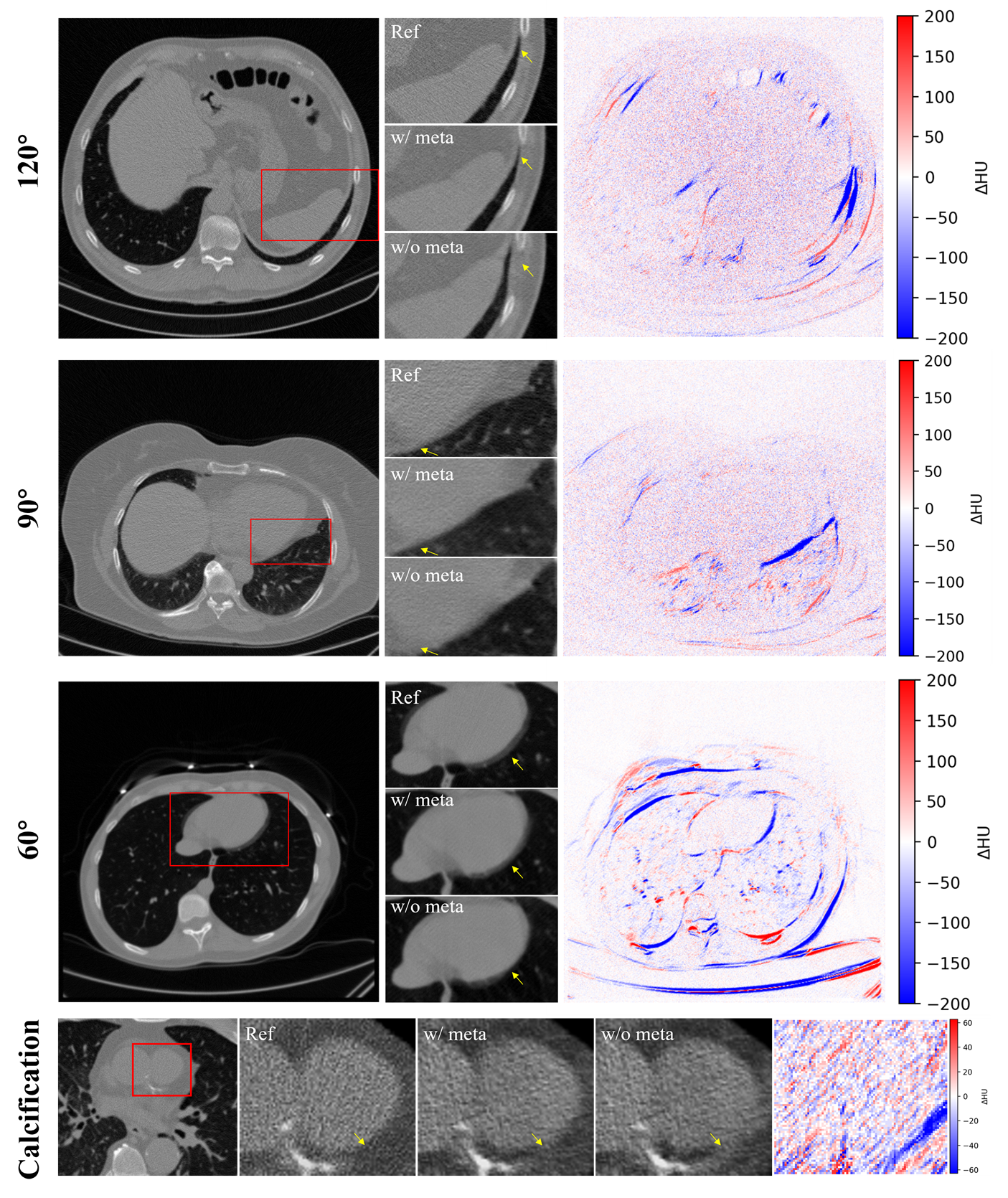}}
\caption{{Task-oriented qualitative ablation of metadata conditioning. Rows 1--3 show representative cases under 120{\degree}, 90{\degree}, and 60{\degree} limited-angle settings, respectively. For each case, we display the full-angle reference (\emph{Ref}), zoomed ROIs (red boxes) for reconstructions \emph{w/ meta} and \emph{w/o meta}, and the corresponding HU absolute-error difference maps computed as $|x_{w/~meta}-x_{Ref}|-|x_{w/o~meta}-x_{Ref}|$. The bottom row shows a vascular calcification zoom-in at 90{\degree}. Arrows indicate reduced streak/bias residuals or vessel-trailing/elongation-like deviations with metadata conditioning.}}
\label{fig9}
\end{figure}

Taken together, the single-category ablations show that each metadata family provides non-redundant information: \emph{Phy} yields the most consistent SSIM benefits, while \emph{Demo} and \emph{Diag} offer complementary global or semantic prior. Removing any one of them weakens reconstruction fidelity, with the effect being more evident under more ill-posed truncation.
{This trend is further supported by the image-domain absolute-error difference maps\cite{li2025diffusion} in Fig. \ref{fig9}. For each truncation setting, we first compute the residual maps for both variants with respect to the full-angle reference, and then visualize the absolute-error difference in HU. Negative values indicate that metadata conditioning reduces the local absolute error compared with the no-metadata variant where the T5X cross-attention receives a null token. Across 120{\degree}, 90{\degree}, and 60{\degree}, metadata conditioning consistently reduces structured streak-like residuals and localized intensity bias, particularly in regions dominated by truncation artifacts (arrows), with more pronounced gains under the more ill-posed 60{\degree} truncation.}

The vascular calcification zoom-in at 90{\degree} (bottom row) provides a clinically relevant example. Compared with the no-metadata variant, removing metadata may lead to spurious vessel trailing/elongation artifacts, whereas metadata conditioning suppresses these patterns and improves anatomical consistency with the full-angle reference.

\subsubsection{{Impacts of the Two-stage Design}}
{We examine the impacts of the two-stage design on CTRATE under 60{\degree}/90{\degree}/120{\degree} truncation by comparing three settings while keeping the network backbone, ADMM/DC configuration, training data, and evaluation protocol unchanged. \textbf{Single Stage (25 steps)} denotes a Stage-II-only pipeline with the standard 25 sampling steps. \textbf{Single Stage (100 steps)} denotes an enhanced single-stage baseline that increases the sampling steps to 100, \emph{i.e.}, a substantially greater sampling steps than the cascade framework. \textbf{Cascaded} denotes the full cascaded framework with 4 + 25 sampling steps (Stage-I + Stage-II). As summarized in Table \ref{tab:ablation}, the cascaded design consistently and substantially outperforms the Stage-II-only baselines across all angles. Relative to \textbf{Single Stage (25 steps)}, the PSNR/SSIM gains reach 7.926 dB/0.201 at 60{\degree}, 12.073 dB/0.201 at 90{\degree}, and 10.779 dB/0.137 at 120{\degree}. Even compared to the step-augmented \textbf{Single Stage (100 steps)}, the proposed approach still improves by 6.988 dB/0.098 at 60{\degree}, 10.664 dB/0.155 at 90{\degree}, and 9.804 dB/0.124 at 120{\degree}, respectively. In addition to higher means, \textbf{Cascaded} also shows lower variability, indicating more stable reconstructions. Overall, these results suggest that performance gains stem from the coarse-to-fine two-stage design rather than simply increasing the sampling steps, where Stage-I supplies a metadata-consistent coarse prior that narrows the solution space and better conditions the subsequent DC-corrected refinement in Stage-II, leading to improved fidelity and stability.}

\subsubsection{{Impacts of SFT}}
{We further evaluate the proposed SFT, where Stage-I and Stage-II are jointly optimized by backpropagating the EDM $v$-prediction loss through the entire cascade. As reported in Table \ref{tab:ablation}, enabling SFT consistently improves reconstruction quality across all truncation angles. Specifically, PSNR/SSIM increases from 34.140/0.916 to 37.072/0.935 at 120{\degree}, from 30.176/0.817 to 33.479/0.888 at 90{\degree}, and from 25.692/0.688 to 29.854/0.801 at 60{\degree}, with the largest gains observed under the most ill-posed 60{\degree} setting. These results indicate that joint fine-tuning effectively reduces cross-stage error accumulation, yielding higher fidelity without increasing test-time requirements.}

\subsubsection{{Impacts of DC Frequency}}
{We study how frequently data consistency should be enforced during diffusion sampling, and whether overly frequent DC diminishes the effects of the learned prior including metadata conditioning. In LACT, the measured sinogram constrains only the observed subspace, whereas missing-wedge (null-space) components must be inferred by the learned prior. As summarized in Table \ref{tab:ablation}, we change the DC schedule while keeping all other factors fixed. Since stage-I only uses 4 sampling steps, DC is applied after each step with a moderate setting, where the weight of DC is 1.0 and the number of ADMM iterations is 10 per DC solver. The ablation varies only the DC frequency in Stage-II with 25 sampling steps. Specifically, we compare w/o DC, DC every step (\emph{E1}), DC every 5 steps (\emph{E5}, default), DC every 10 steps (\emph{E10}), last-step-only DC, and a DC-only baseline without diffusion sampling.}

{The results indicate that DC and the metadata-guided diffusion prior are complementary rather than one dominating the other. First, removing DC substantially degrades performance, \emph{e.g.}, at 90{\degree}, PSNR/SSIM drops from 33.479/0.888 (\emph{E5}) to 30.032/0.716; at 60{\degree}, from 29.854/0.801 (\emph{E5}) to 27.993/0.698 (w/o DC), confirming the importance of DC for enforcing physics fidelity. Second, DC alone is insufficient to achieve high-quality reconstruction, \emph{e.g.}, the DC-only baseline reaches only 30.961/0.810 at 90{\degree} and 27.320/0.688 at 60{\degree}. This indicates that the diffusion model, where metadata enters via conditioning, remains the key driver for recovering missing-angle structures beyond what DC can provide. Third, the DC frequency reveals a clear trade-off: periodic DC consistently outperforms last-step-only DC (\emph{e.g.}, at 60{\degree}, 29.854/0.801 (\emph{E5}) vs 28.531/0.754), suggesting that applying DC intermittently during sampling stabilizes the trajectory without collapsing it to a purely projection-driven solution. Moreover, applying DC at every step (\emph{E1}) offers limited and inconsistent gains over \emph{E5} while increasing computational cost. Based on this ablation, we adopt \emph{E5} as the default DC schedule in Stage-II, achieving near-top performance across angles while reducing the number of DC solvers by 5$\times$ compared with \emph{E1}.}

{We additionally report end-to-end inference runtime under a clearly specified setup. All networks are trained on an NVIDIA A100 GPU, while inference is benchmarked on an RTX 5090 with batch size 1. The timing measures reconstruction only and excludes any I/O. Under the default configuration in Table \ref{tab:ablation}, reconstructing a 2.5D slab of size $512\times512\times3$ takes 47.3 s, which is practical for slice-level inspection or ROI-focused analysis. However, scaling to a full $512\times512\times239$ volume using non-overlapping 3-slice slabs requires 80 slabs and takes approximately 3784 s (63.1 min). This indicates that the current implementation is mainly suited for offline reconstruction and targeted evaluation, and improving throughput with fully 3D inference is an important direction for future work.}

\subsection{{Robustness and Uncertainty}}
\subsubsection{{Representative Failure/Degraded Examples}}
\begin{figure}[!b]
\centerline{\includegraphics[width=3.0in]{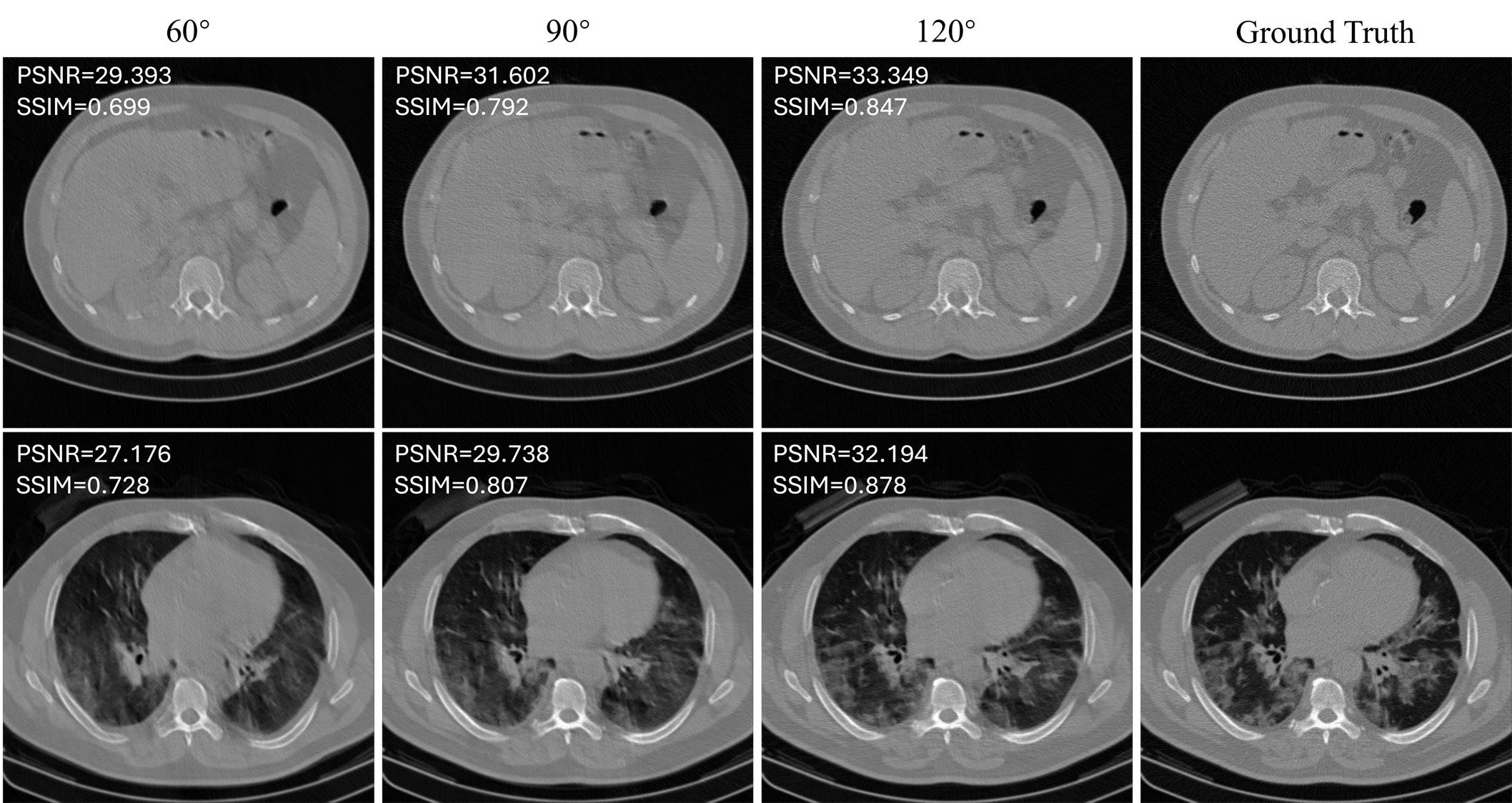}}
\caption{{Representative degraded examples on the CTRATE testing set selected by ranking. The top row shows the lowest-SSIM slice and the bottom row shows the lowest-PSNR slice. PSNR/SSIM are annotated.}}
\label{fig10}
\end{figure}

{To avoid over-representing successful reconstructions, Fig. \ref{fig10} reports worst-ranked slices from the CTRATE testing set. For each acquisition setting (60{\degree}, 90{\degree}, 120{\degree}), we compute slice-wise SSIM and PSNR against the corresponding full-angle reference over all testing slices, and select the lowest-SSIM and lowest-PSNR slices by ranking without manual cherry-picking.
The lowest-SSIM cases typically correspond to low-contrast abdominal slices, where weak soft-tissue contrast makes SSIM sensitive to small intensity bias and residual limited-angle artifacts. In contrast, the lowest-PSNR cases often contain spatially heterogeneous textures and complex pathological patterns, where PSNR penalizes localized deviations around high-frequency textures and lesion boundaries.
Even looking at these worst-ranked slices, the proposed method still achieves the best quantitative performance among the compared methods. On the lowest-SSIM slices (Fig. \ref{fig10}, top row), it achieves PSNR/SSIM of 33.349/0.847 at 120{\degree}, 31.602/0.792 at 90{\degree}, and 29.393/0.699 at 60{\degree}, compared with 32.399/0.827, 28.020/0.746, and 25.725/0.638 from the best competing baseline, respectively. On the lowest-PSNR slices (Fig. \ref{fig10}, bottom row), it achieves 32.194/0.878 at 120{\degree}, 29.738/0.807 at 90{\degree}, and 27.176/0.728 at 60{\degree}, compared with 31.138/0.858, 26.857/0.771, and 24.329/0.672 from the best competing baseline. Nevertheless, due to the intrinsic ill-posedness of limited-angle CT, residual discrepancies and occasional structural deviations remain inevitable under severe information loss, as also observed in Figs.~\ref{fig4}--\ref{fig7}.}
\subsubsection{{Reconstruction Consistency}}
{Beyond mean performance, we quantify slice-level consistency by reporting a relative success-rate criterion. For each acquisition setting, we identify the best competing baseline and then evaluate performance on a per-slice basis. A slice is counted as a "success" if the proposed method achieves SSIM and PSNR values that are both no worse than the best baseline on the same slice. Under this criterion, the proposed method attains success rates of 96.12\% at 120{\degree}, 93.96\% at 90{\degree}, and 84.48\% at 60{\degree}. These results indicate that the proposed method improves the best competing baseline on the majority of slices. The gradual decrease from 120{\degree} to 60{\degree} quantitatively reflects the expected increase in ill-posedness as angular coverage shrinks.}
\subsubsection{{Metadata Perturbation}}
{Metadata plays an important role in our framework by providing semantic conditioning signals to guide the coarse reconstruction in Stage-I and influence the subsequent refinement in Stage-II. In practice, however, metadata may be imperfect, which could potentially generate bias in the learned prior. To evaluate this sensitivity, we conduct a metadata perturbation study by keeping the measured sinogram unchanged while replacing the input metadata with mismatched values. As shown in Table \ref{tab:misinform}, metadata perturbation leads to a consistent yet graceful degradation across 120{\degree}, 90{\degree}, and 60{\degree} settings, where PSNR decreases by 0.642/0.897/1.218 dB and SSIM decreases by 0.009/0.021/0.027, respectively. The same trend is reflected by slight reductions in nMI/PCC and increased RMSE$_\text{HU}$. Overall, these results confirm that accurate metadata is beneficial, while the proposed method remains reasonably stable under metadata inaccuracies, aided by the data-consistency correction that mitigates severe structural drifts.}

{Finally, we also evaluate robustness to dose/noise variations across multiple mAs levels. Due to space, detailed results are provided in Supplementary Table S2 and Fig. S1.}

\begin{table}[t]
\centering
\caption{Study on robustness to metadata perturbation.}
\resizebox{\columnwidth}{!}{
\begin{tabular}{l ccccc}
\toprule
\textbf{Method} & SSIM & PSNR & nMI & PCC & {$\mathrm{RMSE}_{\mathrm{HU}}$} \\
\midrule
\multicolumn{6}{c}{\textbf{120{\degree}}}\\
\midrule
Matched~           & $0.935\pm0.029$ & $37.072\pm3.739$ & $1.358\pm0.038$ & $0.995\pm0.005$ & $23.0\pm5.7$ \\
Mismatched~       & $0.926\pm0.028$ & $36.430\pm3.830$ & $1.345\pm0.038$ & $0.994\pm0.006$ & $25.5\pm5.8$ \\
\midrule
\multicolumn{6}{c}{\textbf{90{\degree}}}\\
\midrule
Matched~           & $0.888\pm0.036$ & $33.479\pm4.081$ & $1.308\pm0.027$ & $0.988\pm0.008$ & $38.2\pm8.5$ \\
Mismatched~       & $0.867\pm0.036$ & $32.582\pm4.157$ & $1.294\pm0.027$ & $0.985\pm0.008$ & $39.5\pm9.1$ \\
\midrule
\multicolumn{6}{c}{\textbf{60{\degree}}}\\
\midrule
Matched~           & $0.801\pm0.053$ & $29.854\pm4.491$ & $1.268\pm0.024$ & $0.969\pm0.015$ & $56.8\pm14.3$\\
Mismatched~       & $0.774\pm0.052$ & $28.636\pm4.523$ & $1.261\pm0.024$ & $0.962\pm0.016$ & $62.3\pm14.7$\\
\bottomrule
\end{tabular}
}
\label{tab:misinform}
\vspace{-2mm}
\end{table}

\section{Discussion and Conclusion}
\label{sec:discussion}
This work introduces a two-stage diffusion framework that integrates structured clinical metadata to guide LACT reconstruction. 
Across simulated and real-world evaluations, the proposed approach improves artifact suppression and anatomical fidelity, and consistently outperforms metadata-free baselines in both quantitative metrics and qualitative assessments.

\begin{figure}[t]
    \centering
    \subfloat[Stage-I: PSNR/SSIM versus sampling steps (1-5).]{%
    \includegraphics[width=\linewidth]{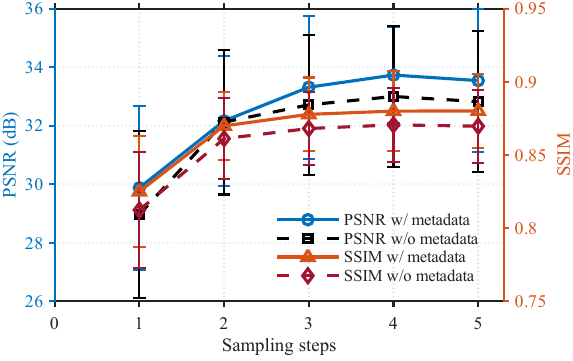}%
    \label{fig11:stageI}%
  }\par\vspace{0.6em}

    \subfloat[Stage-II: PSNR/SSIM versus sampling steps (1-100).]{%
    \includegraphics[width=\linewidth]{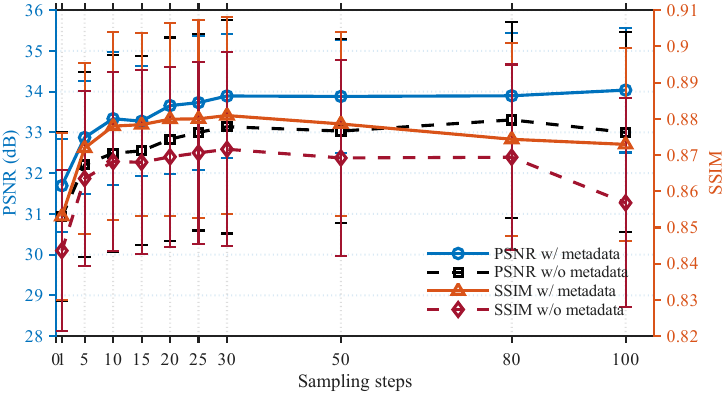}%
    \label{fig11:stageII}%
  }
    \caption{{Effects of sampling-step selection for models w/ and w/o metadata guidance on CTRATE under 90{\degree} limited-angle acquisition. (a) Stage-I with 1 to 5 steps. (b) Stage-II with 1 to 100 steps. PSNR and SSIM are reported with respect to steps.}}
    \label{fig11}
\end{figure}

To guide sampling-step selection for practical deployment, Fig.~\ref{fig11} reports PSNR/SSIM as a function of the sampling-step budget on CTRATE under 90{\degree} acquisition, for both the metadata-guided model and a metadata-free counterpart. In Stage-I (Fig.~\ref{fig11}a), performance improves rapidly from 1 to 3 steps and largely saturates by 4 steps, while additional steps bring negligible gains, supporting our default 4-step coarse stage. In Stage-II (Fig.~\ref{fig11}b), increasing the step budget yields clear improvements up to a moderate range (around 25 to 30 steps), after which the curves become flat and further steps provide limited benefit despite linearly increased runtime. SSIM also shows a mild degradation at large budgets, suggesting diminishing returns and potential over-refinement. Based on this point of diminishing returns, we adopt 25 steps for Stage-II as a favorable fidelity–latency trade-off. Across both stages, metadata guidance provides a consistent but modest quality margin over the metadata-free variant, while the overall step-performance trend remains similar, indicating that step selection is primarily governed by the sampler budget rather than being highly sensitive to the conditioning signal.

We further examine cross-domain generalization, which remains a key challenge for learning-based reconstruction. As observed on external clinical data (Fig. \ref{fig7} (b)), purely supervised baselines may exhibit residual intensity drift or limited-angle artifacts under scanner-domain shift, whereas unsupervised and weakly-supervised methods that explicitly incorporate data-consistency constraints are generally more robust. {In addition, our method achieves excellent reconstruction quality under unseen fan-beam and parallel-beam geometries as shown in Table \ref{tab:quantitative_2} and Figs.~\ref{fig5}--\ref{fig6}. This robustness stems from the decoupled design between the semantic diffusion prior and the physics module. During inference, the forward operator $A$ and its adjoint $A^T$ in the ADMM module are instantiated via CTLIB to match the testing geometry, enforcing projection-level fidelity for the given detector spacing, distances and angular coverage, while the diffusion prior remains geometry-agnostic and is further stabilized by metadata conditioning. Nevertheless, residual domain mismatch can still manifest, especially in scanner- and protocol-dependent background structures (\emph{e.g.}, the patient table), motivating additional calibration beyond standard training. In the future, we will therefore explore post-training target-domain adaptation of the diffusion prior, such as Group Relative Policy Optimization (GRPO)\cite{shao2024deepseekmath} style reward optimization driven by physics-based consistency and artifact-aware objectives, to further improve robustness under scanner and protocol shifts.}

\begin{figure*}[!t]
\centerline{\includegraphics[width=6.9in]{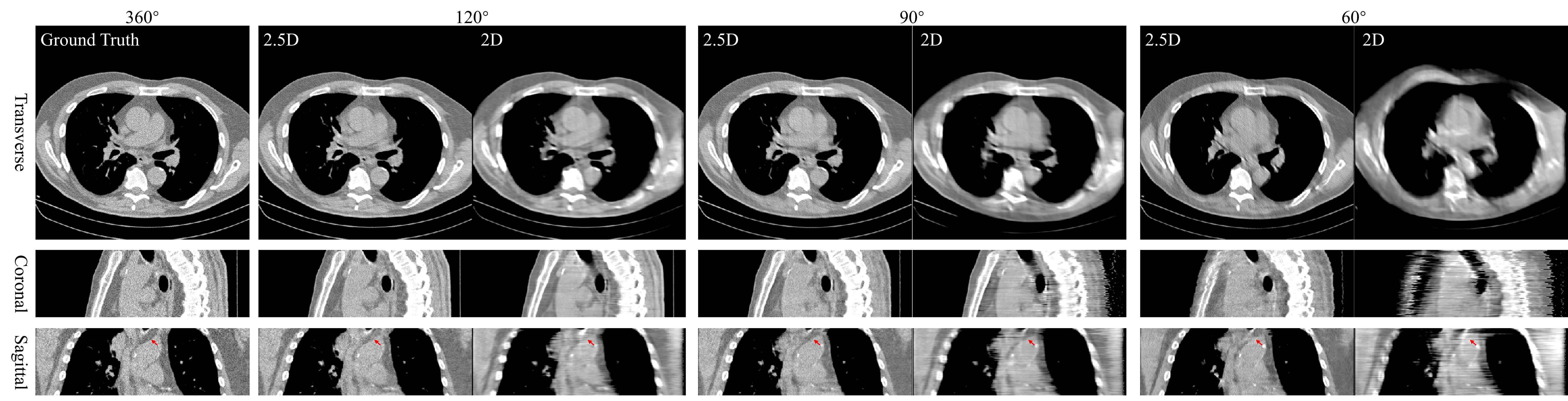}}
\caption{{Qualitative comparison between 2.5D and 2D variants to illustrate axial-context loss. The arrows highlight regions where the 2D model weakens the axial visibility or contrast of elongated structures in reformatted views. The display window is $[-400, 250]$ HU}}
\label{fig12}
\end{figure*}
{Beyond robustness to domain and geometry shifts, another practical consideration for application is preserving through-plane anatomical continuity when the reconstructed volume is inspected using multiplanar reformation (MPR).
A limitation of purely slice-wise 2D learning is the loss of short-range axial context, which can weaken the visibility, contrast and continuity of elongated structures in coronal and sagittal planes\cite{xu2025Swapnet}. To mitigate this issue, we adopt a lightweight 2.5D slab-based design: Stage-I predicts a coarse 3D slab of 9 adjacent axial slices, and Stage-II performs 2.5D diffusion refinement by conditioning on the 9-slice slab while reconstructing the central three slices. As illustrated in Fig. \ref{fig12}, the 2.5D variant better preserves cross-slice continuity, whereas the 2D variant tends to attenuate the axial visibility and contrast of elongated structures in reformatted views highlighted by arrows. The main reason for not pursuing a full volumetric diffusion model at clinical resolution is feasibility: volumetric diffusion together with the ADMM-based data-consistency correction would significantly increase memory and computation cost. Under the same batch size of 2 and image resolution, peak training memory increases from 20.9 GB (2D) to 29.4 GB (2.5D) with fine-tuning enabled, whereas a full 3D variant would likely exceed practical single-GPU limits. Overall, the slab-based 2.5D design injects short-range axial context to improve through-plane continuity, without the prohibitive cost of full 3D diffusion.
A full volumetric model remains a promising direction as more efficient 3D backbones and computational resources become available.}

Beyond the architectural trade-offs discussed above, a remaining limitation is the granularity and localization of diagnostic metadata. In our framework, diagnostic report is inherently scan-level and thus shared across slices, which may be suboptimal for highly localized findings and slice-dependent variations. Recent work\cite{li2026} shows that fine-grained processing of diagnostic text, such as extracting concise, localized, and disease-relevant phrases from radiology reports, can further enhance text-guided image generation. Inspired by this, future work will improve the semantic conditioning pipeline by bridging textual and spatial granularity through slice-aware report alignment and refined extraction of diagnostic descriptors.

In conclusion, we propose a metadata-guided two-stage diffusion framework for LACT reconstruction that improves reconstruction fidelity under severe truncation while enhancing sampling efficiency and cross-domain robustness. The incorporation of structured clinical priors proves a simple yet effective mechanism for constraining the reconstruction problem under missing-angle measurements. Future work will focus on finer-grained semantic alignment, accelerating inference, and improving robustness and scalability via target-domain post-training adaptation to facilitate broader clinical deployment.





\FloatBarrier
\bibliographystyle{IEEEtran}  
\bibliography{ref}  

\begin{thebibliography}{10}
\providecommand{\url}[1]{#1}
\csname url@samestyle\endcsname
\providecommand{\newblock}{\relax}
\providecommand{\bibinfo}[2]{#2}
\providecommand{\BIBentrySTDinterwordspacing}{\spaceskip=0pt\relax}
\providecommand{\BIBentryALTinterwordstretchfactor}{4}
\providecommand{\BIBentryALTinterwordspacing}{\spaceskip=\fontdimen2\font plus
\BIBentryALTinterwordstretchfactor\fontdimen3\font minus \fontdimen4\font\relax}
\providecommand{\BIBforeignlanguage}[2]{{%
\expandafter\ifx\csname l@#1\endcsname\relax
\typeout{** WARNING: IEEEtran.bst: No hyphenation pattern has been}%
\typeout{** loaded for the language `#1'. Using the pattern for}%
\typeout{** the default language instead.}%
\else
\language=\csname l@#1\endcsname
\fi
#2}}
\providecommand{\BIBdecl}{\relax}
\BIBdecl

\bibitem{lubbers2018comprehensive}
M.~Lubbers \emph{et~al.}, ``Comprehensive cardiac {CT} with myocardial perfusion imaging versus functional testing in suspected coronary artery disease: the multicenter, randomized {CRESCENT-II} trial,'' \emph{JACC: Cardiovascular Imaging}, vol.~11, no.~11, pp. 1625--1636, 2018.

\bibitem{national2011reduced}
{National Lung Screening Trial Research Team}, ``Reduced lung-cancer mortality with low-dose computed tomographic screening,'' \emph{New England Journal of Medicine}, vol. 365, no.~5, pp. 395--409, 2011.

\bibitem{de2020reduced}
H.~J. de~Koning \emph{et~al.}, ``Reduced lung-cancer mortality with volume {CT} screening in a randomized trial,'' \emph{New England journal of medicine}, vol. 382, no.~6, pp. 503--513, 2020.

\bibitem{he2025three}
H.~He, C.~Yu, Y.~Yang, J.~G. Maessen, and P.~Sardari~Nia, ``Three-dimensional reconstruction and virtual simulation of patient-specific anatomy for procedural planning in thoracoscopic segmentectomy: A systematic review and meta-analysis,'' \emph{European Journal of Cardio-Thoracic Surgery}, vol.~67, no.~9, p. ezaf283, 2025.

\bibitem{key2023cone}
B.~M. Key, S.~M. Tutton, and M.~J. Scheidt, ``Cone-beam {CT} with enhanced needle guidance and augmented fluoroscopy overlay: applications in interventional radiology,'' \emph{American Journal of Roentgenology}, vol. 221, no.~1, pp. 1--10, 2023.

\bibitem{Kashyap_2025}
M.~Kashyap \emph{et~al.}, ``Automated deep learning–based detection and segmentation of lung tumors at {CT} imaging,'' \emph{Radiology}, vol. 314, no.~1, p. e233029, 2025.

\bibitem{LEE2024169393}
M.~Lee \emph{et~al.}, ``Development of learning-based online c-arm {CT} imaging technique for image-guided adaptive brachytherapy,'' \emph{Nuclear Instruments and Methods in Physics Research Section A: Accelerators, Spectrometers, Detectors and Associated Equipment}, vol. 1064, p. 169393, 2024.

\bibitem{toia2020technical}
P.~Toia \emph{et~al.}, ``Technical development in cardiac {CT}: current standards and future improvements—a narrative review,'' \emph{Cardiovascular diagnosis and therapy}, vol.~10, no.~6, p. 2018, 2020.

\bibitem{mergen2023importance}
V.~Mergen \emph{et~al.}, ``The importance of temporal resolution for ultra-high-resolution coronary angiography: evidence from photon-counting detector {CT},'' \emph{Investigative Radiology}, vol.~58, no.~11, pp. 767--774, 2023.

\bibitem{sartoretti2025effect}
T.~Sartoretti \emph{et~al.}, ``Effect of temporal resolution on calcium scoring: insights from photon-counting detector {CT},'' \emph{The International Journal of Cardiovascular Imaging}, vol.~41, no.~3, pp. 615--625, 2025.

\bibitem{Zhou2021limited}
B.~Zhou, S.~K. Zhou, J.~S. Duncan, and C.~Liu, ``Limited view tomographic reconstruction using a cascaded residual dense spatial-channel attention network with projection data fidelity layer,'' \emph{IEEE Transactions on Medical Imaging}, vol.~40, no.~7, pp. 1792--1804, 2021.

\bibitem{apfaltrer2013enhanced}
P.~Apfaltrer \emph{et~al.}, ``Enhanced temporal resolution at cardiac {CT} with a novel {CT} image reconstruction algorithm: initial patient experience,'' \emph{European Journal of Radiology}, vol.~82, no.~2, pp. 270--274, 2013.

\bibitem{bappy2025deep}
D.~Bappy, D.~Kang, J.~Lee, Y.~Lee, and H.~Baek, ``Deep prior based limited-angle tomography,'' in \emph{International Conference on Pattern Recognition (ICPR)}.\hskip 1em plus 0.5em minus 0.4em\relax Springer, 2025, pp. 79--95.

\bibitem{ZHANG2021102030}
Z.~Zhang, B.~Chen, D.~Xia, E.~Y. Sidky, and X.~Pan, ``{Directional-TV} algorithm for image reconstruction from limited-angular-range data,'' \emph{Medical Image Analysis}, vol.~70, p. 102030, 2021.

\bibitem{WANG2023128013}
C.~Wang, X.~Wang, K.~Zhao, M.~Huang, X.~Li, and W.~Yu, ``A cascading l0 regularization reconstruction method in nonsubsampled contourlet domain for limited-angle {CT},'' \emph{Applied Mathematics and Computation}, vol. 451, p. 128013, 2023.

\bibitem{huang2018scale}
Y.~Huang, O.~Taubmann, X.~Huang, V.~Haase, G.~Lauritsch, and A.~Maier, ``Scale-space anisotropic total variation for limited angle tomography,'' \emph{IEEE Transactions on Radiation and Plasma Medical Sciences}, vol.~2, no.~4, pp. 307--314, 2018.

\bibitem{xu2024hybrid}
Y.~Xu, S.~Han, D.~Wang, G.~Wang, J.~S. Maltz, and H.~Yu, ``Hybrid u-net and swin-transformer network for limited-angle cardiac computed tomography,'' \emph{Physics in Medicine \& Biology}, vol.~69, no.~10, p. 105012, 2024.

\bibitem{Jin2017}
K.~H. Jin, M.~T. McCann, E.~Froustey, and M.~Unser, ``Deep convolutional neural network for inverse problems in imaging,'' \emph{IEEE Trans. Imag. Process.}, vol.~26, no.~9, pp. 4509--4522, 2017.

\bibitem{pan2022multi}
J.~Pan, H.~Zhang, W.~Wu, Z.~Gao, and W.~Wu, ``Multi-domain integrative swin transformer network for sparse-view tomographic reconstruction,'' \emph{Patterns}, vol.~3, no.~6, 2022.

\bibitem{Bahareh2025}
B.~Morovati, S.~Han, L.~Zhou, D.~Wang, and H.~Yu, ``Photon-counting {CT} reconstruction using separable attention-based tensor neural network prior,'' in \emph{2025 IEEE 22nd International Symposium on Biomedical Imaging (ISBI)}, 2025, pp. 1--4.

\bibitem{ma2023prompted}
C.~Ma, Z.~Li, J.~He, J.~Zhang, Y.~Zhang, and H.~Shan, ``Prompted contextual transformer for incomplete-view {CT} reconstruction,'' \emph{arXiv preprint arXiv:2312.07846}, 2023.

\bibitem{ma2025tmi}
C.~Ma \emph{et~al.}, ``Radiologist-in-the-loop self-training for generalizable {CT} metal artifact reduction,'' \emph{IEEE Trans. Med. Imag.}, vol.~44, no.~6, pp. 2504--2514, 2025.

\bibitem{li2024progressively}
J.~Li, W.~Du, H.~Cui, H.~Chen, Y.~Zhang, and H.~Yang, ``Progressively prompt-guided models for sparse-view {CT} reconstruction,'' \emph{IEEE Transactions on Radiation and Plasma Medical Sciences}, vol.~9, no.~4, pp. 447--459, 2025.

\bibitem{shi2026prompt}
B.~Shi, B.~Chen, S.~Zhang, H.~Fu, and Z.~Hu, ``Prompt guiding multi-scale adaptive sparse representation-driven network for low-dose {CT MAR},'' \emph{Medical Image Analysis}, vol. 108, p. 103870, 2026.

\bibitem{gauriau2020using}
R.~Gauriau \emph{et~al.}, ``Using {DICOM} metadata for radiological image series categorization: a feasibility study on large clinical brain {MRI} datasets,'' \emph{Journal of Digital Imaging}, vol.~33, no.~3, pp. 747--762, 2020.

\bibitem{bao2023all}
F.~Bao \emph{et~al.}, ``All are worth words: A vit backbone for diffusion models,'' in \emph{Proceedings of the IEEE/CVF Conference on Computer Vision and Pattern Recognition}, 2023, pp. 22\,669--22\,679.

\bibitem{ma2024efficient}
Z.~Ma \emph{et~al.}, ``Efficient diffusion models: A comprehensive survey from principles to practices,'' \emph{arXiv preprint arXiv:2410.11795}, 2024.

\bibitem{chung2025contextmri}
H.~Chung, D.~Lee, Z.~Wu, B.-H. Kim, K.~L. Bouman, and J.~C. Ye, ``{ContextMRI}: Enhancing compressed sensing {MRI} through metadata conditioning,'' \emph{arXiv preprint arXiv:2501.04284}, 2025.

\bibitem{Hamamci2024GenerateCT}
I.~E. Hamamci \emph{et~al.}, ``{GenerateCT}: Text-conditional generation of {3D} chest {CT} volumes,'' in \emph{European Conference on Computer Vision}, 2024, pp. 126--143.

\bibitem{raffel2020exploring}
C.~Raffel \emph{et~al.}, ``Exploring the limits of transfer learning with a unified text-to-text transformer,'' \emph{Journal of Machine Learning Research}, vol.~21, no. 140, pp. 1--67, 2020.

\bibitem{roberts2022t5x}
A.~Roberts \emph{et~al.}, ``Scaling up models and data with $\texttt{t5x}$ and $\texttt{seqio}$,'' \emph{arXiv preprint arXiv:2203.17189}, 2022.

\bibitem{hamamci2024developing}
I.~E. Hamamci \emph{et~al.}, ``Developing generalist foundation models from a multimodal dataset for {3D} computed tomography,'' \emph{arXiv preprint arXiv:2403.17834}, 2024.

\bibitem{xia2021magic}
W.~Xia \emph{et~al.}, ``Magic: Manifold and graph integrative convolutional network for low-dose {CT} reconstruction,'' \emph{IEEE Trans. Med. Imag.}, vol.~40, no.~12, pp. 3459--3472, 2021.

\bibitem{Liu_2012}
Y.~Liu, J.~Ma, Y.~Fan, and Z.~Liang, ``Adaptive-weighted total variation minimization for sparse data toward low-dose x-ray computed tomography image reconstruction,'' \emph{Physics in Medicine \& Biology}, vol.~57, no.~23, p. 7923, 2012.

\bibitem{chung2023decomposed}
H.~Chung, S.~Lee, and J.~C. Ye, ``Decomposed diffusion sampler for accelerating large-scale inverse problems,'' \emph{arXiv preprint arXiv:2303.05754}, 2023.

\bibitem{gao2018combined}
R.~Gao, F.~Tronarp, and S.~S{\"a}rkk{\"a}, ``Combined {analysis-L1} and total variation {ADMM} with applications to meg brain imaging and signal reconstruction,'' in \emph{2018 26th European Signal Processing Conference (EUSIPCO)}.\hskip 1em plus 0.5em minus 0.4em\relax IEEE, 2018, pp. 1930--1934.

\bibitem{jin2017deep}
K.~H. Jin, M.~T. McCann, E.~Froustey, and M.~Unser, ``Deep convolutional neural network for inverse problems in imaging,'' \emph{IEEE Trans. Imag. Process.}, vol.~26, no.~9, pp. 4509--4522, 2017.

\bibitem{liu2023dolce}
J.~Liu \emph{et~al.}, ``Dolce: A model-based probabilistic diffusion framework for limited-angle {CT} reconstruction,'' in \emph{Proceedings of the IEEE/CVF International Conference on Computer Vision}, 2023, pp. 10\,498--10\,508.

\bibitem{studholme1999overlap}
C.~Studholme, D.~L. Hill, and D.~J. Hawkes, ``An overlap invariant entropy measure of {3D} medical image alignment,'' \emph{Pattern Recognition}, vol.~32, no.~1, pp. 71--86, 1999.

\bibitem{AIoa2400937}
R.~Hagopian \emph{et~al.}, ``{AI} opportunistic coronary calcium screening at veterans affairs hospitals,'' \emph{NEJM AI}, vol.~2, no.~6, p. AIoa2400937, 2025.

\bibitem{Pletcher2004}
M.~J. Pletcher, J.~A. Tice, M.~Pignone, and W.~S. Browner, ``Using the coronary artery calcium score to predict coronary heart disease events: A systematic review and meta-analysis,'' \emph{Archives of Internal Medicine}, vol. 164, no.~12, pp. 1285--1292, 2004.

\bibitem{li2025diffusion}
M.~Li, X.~Li, S.~Safai, A.~J. Lomax, and Y.~Zhang, ``Diffusion schr{\"o}dinger bridge models for high-quality {MR-to-CT} synthesis for proton treatment planning,'' \emph{Medical Physics}, vol.~52, no.~7, p. e17898, 2025.

\bibitem{shao2024deepseekmath}
Z.~Shao \emph{et~al.}, ``Deepseekmath: Pushing the limits of mathematical reasoning in open language models,'' \emph{arXiv preprint arXiv:2402.03300}, 2024.

\bibitem{xu2025Swapnet}
X.~Xu, M.~L. Klasky, M.~T. McCann, J.~Hu, and J.~A. Fessler, ``Swap-net: A memory-efficient {2.5D} network for sparse-view {3D} cone beam {CT} reconstruction to {ICF} applications,'' \emph{IEEE Transactions on Computational Imaging}, vol.~11, pp. 872--887, 2025.

\bibitem{li2026}
X.~Li \emph{et~al.}, ``Text-driven tumor synthesis,'' \emph{IEEE Transactions on Medical Imaging}, pp. 1--1, 2026.

\end{thebibliography}

\end{document}